\documentclass[10pt,journal,compsoc]{IEEEtran}
\usepackage{amsmath}
\usepackage{bbding}
\usepackage{amssymb}
\DeclareMathOperator*{\argmin}{argmin}
\usepackage{epsfig}
\usepackage{graphicx}
\usepackage[utf8]{inputenc}
\usepackage[T1]{fontenc}
\usepackage{url}
\usepackage{nicefrac}
\usepackage{microtype}
\usepackage{lipsum}
\usepackage{graphicx}
\usepackage{ragged2e}
\usepackage{booktabs}
\usepackage{amsfonts}
\usepackage{verbatim}
\usepackage{makecell}
\usepackage{multirow}

\usepackage{listings}
\usepackage{color}
\usepackage[dvipsnames]{xcolor}
\definecolor{dkgreen}{rgb}{0,0.6,0}
\definecolor{gray}{rgb}{0.5,0.5,0.5}
\definecolor{mauve}{rgb}{0.58,0,0.82}
\usepackage{pifont}
\usepackage{bm}
\usepackage{caption}
\usepackage{booktabs,subcaption,amsfonts,dcolumn}
\usepackage[ruled,vlined]{algorithm2e}
\usepackage[switch]{lineno}
\usepackage[pagebackref=false,breaklinks=false,colorlinks,citecolor=brown,bookmarks=false]{hyperref}

\usepackage{fancyhdr}
\usepackage{colortbl}
\usepackage{color}
\usepackage[dvipsnames]{xcolor}
\definecolor{mygray}{gray}{.93}
\begin{document}
\title{PASS++: A Dual Bias Reduction Framework for Non-Exemplar Class-Incremental Learning}

\author{Fei Zhu,
	Xu-Yao Zhang,
	Zhen Cheng,
	Cheng-Lin Liu~\IEEEmembership{Fellow,~IEEE}
	\IEEEcompsocitemizethanks{
		\IEEEcompsocthanksitem Fei Zhu is with the Centre for Artificial Intelligence and Robotics, Hong Kong Institute of Science and Innovation, Chinese Academy of Sciences, Hong Kong 999077, China.
		\IEEEcompsocthanksitem Xu-Yao Zhang, Zhen Cheng and Cheng-Lin Liu are with the State Key Laboratory of Multimodal Artificial Intelligence Systems, Institute of Automation of Chinese Academy of Sciences, 95 Zhongguancun East Road, Beijing 100190, P.R. China, and also with the School of Artificial Intelligence, University of Chinese Academy of Sciences, Beijing 100049, P.R. China.
		\IEEEcompsocthanksitem Email: \{zhufei2018, chengzhen2019\}@ia.ac.cn, \{xyz, liucl\}@nlpr.ia.ac.cn}
}

\IEEEtitleabstractindextext{
\begin{abstract}
   Class-incremental learning (CIL) aims to recognize new classes incrementally while maintaining the discriminability of old classes. Most existing CIL methods are exemplar-based, i.e., storing a part of old data for retraining. Without relearning old data, those methods suffer from catastrophic forgetting. In this paper, we figure out two inherent problems in CIL, \emph{i.e.}, representation bias and classifier bias, that cause catastrophic forgetting of old knowledge. To address these two biases, we present a simple and novel dual bias reduction framework that employs self-supervised transformation (SST) in input space and prototype augmentation (protoAug) in deep feature space. On the one hand, SST alleviates the representation bias by learning generic and diverse representations that can transfer across different tasks. On the other hand, protoAug overcomes the classifier bias by explicitly or implicitly augmenting prototypes of old classes in the deep feature space, which poses tighter constraints to maintain previously learned decision boundaries. We further propose hardness-aware prototype augmentation and multi-view ensemble strategies, leading to significant improvements. The proposed framework can be easily integrated with pre-trained models.
   Without storing any samples of old classes, our method can perform comparably with state-of-the-art exemplar-based approaches which store plenty of old data. 
   We hope to draw the attention of researchers back to non-exemplar CIL by rethinking the necessity of storing old samples in CIL.

\end{abstract}

}

\maketitle
\IEEEdisplaynontitleabstractindextext
\IEEEpeerreviewmaketitle
\IEEEraisesectionheading{\section{Introduction}\label{sec:introduction}}

\pagestyle{fancy}
\cfoot{}
\rhead{\thepage}
\renewcommand{\headrulewidth}{0pt}
\renewcommand{\footrulewidth}{0pt}

\IEEEPARstart{H}{umans} and large primates have an inherent and distinctive ability to incrementally acquire novel experience and accumulate
knowledge throughout their lifespan. This ability, referred to as incremental learning (continual learning or lifelong learning) \cite{wixted2004psychology, tulving197212}, is crucial for artificial intelligence because training examples in real-world applications usually appear sequentially. For instance, a robot with a set of default object recognition capabilities may encounter new objects that have to be identified in dynamic and open environments. Unfortunately, deep neural networks only perform well on homogenized, balanced, and shuffled data \cite{hadsell2020embracing}, they fail completely or suffer from drastic performance degradation of previously learned tasks after learning new knowledge, which is a well-documented phenomenon known as catastrophic forgetting \cite{Goodfellow2014AnEI, McCloskey1989CatastrophicII}. Therefore, developing models that can learn and accumulate sequential experience is a fundamental but challenging step toward human-level intelligence.

In the past several years, a multitude of works has emerged to enable deep neural networks to preserve and extend knowledge continually. Early studies \cite{Kirkpatrick2017OvercomingCF, Li2018LearningWF, Zenke2017ContinualLT} often consider task-incremental learning (TIL), in which a model continually learns a sequence of distinct tasks, and the task identity is provided at inference time. In this paper, we consider a more realistic and challenging scenario of class-incremental learning (CIL) \cite{masana2022class, wang2024comprehensive, zhou2023deep, Zhu_2021_CVPR, douillard2022dytox}, in which each task in the sequence contains a set of classes disjoint from old classes, and the model needs to learn a unified classifier that can classify all classes seen at different stages.

\begin{figure*}[t]
	\begin{center}
		\centerline{\includegraphics[width=0.93\textwidth]{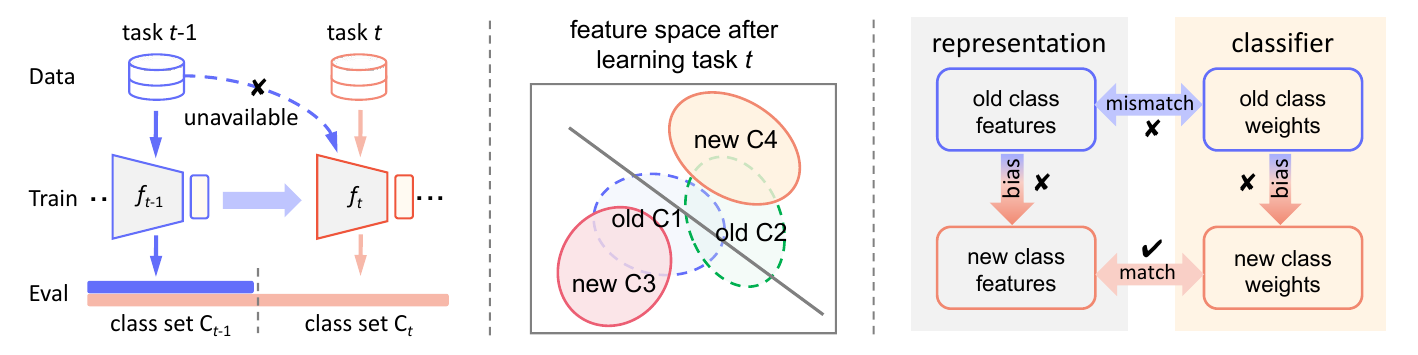}}
		\vskip -0.05in
		\caption{Left: In non-exemplar CIL, the model is updated continuously on new classes without accessing old data. Middle: after learning new classes, previous learned decision boundaries are distorted and the representation confusion is severe. Right: illustration of representation and classifier bias between old and new classes.}
		\label{figure-1}
	\end{center}
	\vskip -0.2in
\end{figure*}

We first demonstrate the challenge of CIL by comparing it with the traditional supervised learning paradigm. For a common supervised classification task, the training data is presented interleaved and used to train a model in an end-to-end manner. Therefore, both the feature extractor and classifier are balanced. Nevertheless, without accessing old data, the learning paradigm of CIL will lead to two inherent problems: \textbf{\textit{representation bias}} and \textbf{\textit{classifier bias}}, as shown in Fig.~\ref{figure-1}. First, for representation learning, if the feature extractor is fixed after learning the old classes, the model could maintain previously learned representations, but lacks plasticity for new classes; on the contrary, if we update the feature extractor on new classes, the old representations could be easily forgotten. We denote this as representation bias. Second, without accessing old training data, the decision boundaries between old and new classes are hard to establish and the classifier will be severely biased toward new classes. We denote this as classifier bias. Consequently, with those two biases, inputs from old classes can be easily predicted as new classes.

To address both the representation and classifier bias in CIL, \textbf{\textit{exemplar-based}} \cite{Rebuffi2017iCaRLIC, Douillard2020SmallTaskIL, liu2021adaptive, yan2021dynamically, wang2022foster, douillard2022dytox, nie2023bilateral, wen2023optimizing, lin2023class, kim2023learnability} methods store a fraction of old data to jointly train the model with current data. With retraining stored data, the representation bias and classifier bias can be largely alleviated. However, storing data is undesirable for a number of reasons. First, it is inefficient to retrain the old data and also hard to scale up for large-scale CIL due to the limited memory resource \cite{parisi2019continual}. Second, storing data may be not allowed in some situations because of privacy and safety concerns \cite{li2020federated}. Moreover, it is less human-like to directly store raw data for memory from a biological perspective \cite{kumaran2016learning}. An alternative way is to learn deep generative models to generate pseudo-samples of old classes
\cite{Shin2017ContinualLW, Wu2018MemoryRG, gao2022r}. Nevertheless, generative models are inefficient to train and also suffer from catastrophic forgetting \cite{Wu2018MemoryRG}. Without data replaying, regularization based methods \cite{Kirkpatrick2017OvercomingCF, lin2022towards, wang2021training, kong2022balancing, lintrgp} address the catastrophic forgetting by identifying and penalizing future changes to some important parameters of the original model. However, they show poor performance in CIL scenario \cite{van2022three}. 
From those works, it seems that storing old samples is a must for CIL. Therefore, a natural question arises:
\textit{can we achieve acceptable CIL performance in a simple and efficient way without storing any old samples?} In this work, we propose a simple and effective dual bias reduction framework for \textbf{\textit{non-exemplar}} (\emph{i.e.}, without storing training data of old classes) CIL by overcoming representation and classifier bias.

For representation learning, we hypothesize that diverse and transferable representations is an important requirement in incremental learning. Considering a simple example of binary classification between sofa and chair, the classes can be separated by only learning the discriminative characteristics (e.g., leg). The ``leg'' feature, however, will
no longer be useful and eventually get forgotten if future task is to classify another
set of images as chair or desk; On the other
hand, if the model learns more transferable and diverse representations like shape and texture, those features may be re-used for
future tasks and remain unforgotten. 
Inspired by the natural connection between incremental learning and self-supervised learning, we propose to leverage the rotation based self-supervised transformation (SST) to reduce the representation bias, revealing surprising yet intriguing findings that SST can boost the performance of CIL significantly.

For classifier learning, we propose to memorize \emph{one} class-representative prototype for each old class. When learning new classes, we generate pseudo-features of old classes by augmenting the memorized prototypes. Then, the pseudo-features are jointly classified with features of new classes to maintain discrimination and balance between old and new classes. Particularly, the above explicit augmentation can be formulated in an implicit way that can theoretically generate an infinite number of features of old classes fast and elegantly.
Furthermore, we present a hardness-aware prototype augmentation strategy to enhance the performance. Compared with the commonly used data replay strategy, protoAug is more memory-efficient and privacy-friendly.

In conclusion, we reveal that the representation bias and classifier bias are crucial for catastrophic forgetting, and propose a simple and effective dual bias reduction framework for non-exemplar CIL. With SST for new classes in input space and prototype augmentation for old classes in deep feature space, our framework addresses the representation bias and classifier bias in a simple and effective manner. Note that if one of those two biases exists, the learned representations would be mismatched with the classifier, as illustrated in Fig.~\ref{figure-1} (Right). Without storing old training data, our method can obtain comparable results with state-of-the-art (SoTA) exemplar-based methods. Parts of this work were published originally in CVPR \cite{Zhu_2021_CVPR}. This paper
extends our earlier works in several important aspects:
\begin{itemize}
	\item Hardness-aware protoAug. We analyze the shortcomings of the original protoAug in detail, and propose a new strategy to synthesize hardness-aware old feature instances by ranking and mixing the feature instances of new classes and old prototypes.
	\item Multi-view ensemble. Considering the classes generated via SST represent meaningful multi-view information, the predictions on the original and transformation classes are fused to boost the performance. The newly introduced hardness-aware protoAug and multi-view ensemble lead to significant performance improvements over our conference version \cite{Zhu_2021_CVPR}.
	\item We extend the proposed framework to pre-trained backbones based on low-rank adaptation \cite{hulora}, and demonstrate that our method can also yield SoTA results among non-exemplar methods and its performance is comparable to the latest exemplar-based approach \cite{lin2023class}.
	\item We conduct experimental comparison with many recent approaches such as distillation-based, dynamic architecture-based and generative exemplar-based methods. We also add large-scale experiments on ImageNet-Full to demonstrate the scalability and generalizability of our method. 
	\item To reflect the robustness of incremental learners, we further propose to evaluate incremental model under distribution shift (e.g., noise, blur, and weather), and provide insightful observations. Additional results, analysis and discussion are presented in Sec. \ref{sec:analysis}.
\end{itemize}

\section{Related Work} \label{sec:related-work}

\subsection{Exemplar-based Class-Incremental Learning}
Although a basic assumption in CIL is the inaccessibility of old training data, exemplar-based methods relax such constraints by saving and retraining a portion of data of old classes. Existing exemplar-based methods can be grouped into three main categories: imbalance calibration, structured distillation, and dynamic architecture.

\vspace*{3pt}
\noindent\textbf{Imbalance calibration} based methods were proposed to alleviate the imbalance problem between new and old classes. Some approaches \cite{castro2018end, Hou2019LearningAU, ahn2021ss, lee2019overcoming, zhou2021co, shi2022mimicking, tang2022learning, kang2022class} aim to learn balanced classifier during training, while another class of approaches \cite{Rebuffi2017iCaRLIC, Wu2019LargeSI, belouadah2019il2m, ZhaoXGZX20, belouadah2020scail, slim2022dataset} use post-hoc methods to calibrate the classifier. ITO \cite{zhu2023imitating} addresses the imbalance problem via weight calibration and feature calibration. RMM \cite{liu2021rmm} uses a reinforcement learning based strategy to learn an optimal memory management policy. Besides, generative exemplar-based methods \cite{Shin2017ContinualLW, smith2021always, KemkerK18, gao2022r, gao2023ddgr} adopt generative model to generate old pseudo-examples, which avoid saving real data and suffer from less imbalance problem.

\vspace*{3pt}
\noindent\textbf{Structured distillation} based methods explore the following question: what kind of knowledge should be distilled to maintain the performance on old classes? Besides unstructured distillation \cite{Rebuffi2017iCaRLIC, Hou2019LearningAU, castro2018end, Douillard2020SmallTaskIL, boschini2022class}, recent works \cite{tao2020topology, simon2021learning, hu2021distilling, cha2021co2l, ashok2022class} focus on structured distillation. For example, TPCIL \cite{tao2020topology} constrains the neighborhood relationships. GeoDL \cite{simon2021learning} distillates features outputted by old and current model in the continuous and infinite intermediate subspaces.

\vspace*{3pt}
\noindent\textbf{Architecture} based methods \cite{rusu2016progressive, mallya2018packnet, kim2021split, liu2021adaptive, pham2021dualnet, yan2021dynamically, wang2022foster, douillard2022dytox, gao2023dkt} focus on the network structure during the course of incremental learning. Specifically, to prevent the forgetting of old tasks, the old parameters are frozen and new branches are allocated to learn new tasks. Among them, early works \cite{rusu2016progressive, mallya2018packnet} require task identity at inference time to identify the corresponding sub-network for each task, which is impractical for CIL. Recently, DER \cite{yan2021dynamically}, FOSTER \cite{wang2022foster}, Dytox \cite{douillard2022dytox}, DKT \cite{gao2023dkt}, and DNE \cite{hu2023dense} design novel architecture to preserve old knowledge while learning new concepts, which yield strong CIL performance.

\begin{figure*}[t]
	\begin{center}
		\centerline{\includegraphics[width=0.85\textwidth]{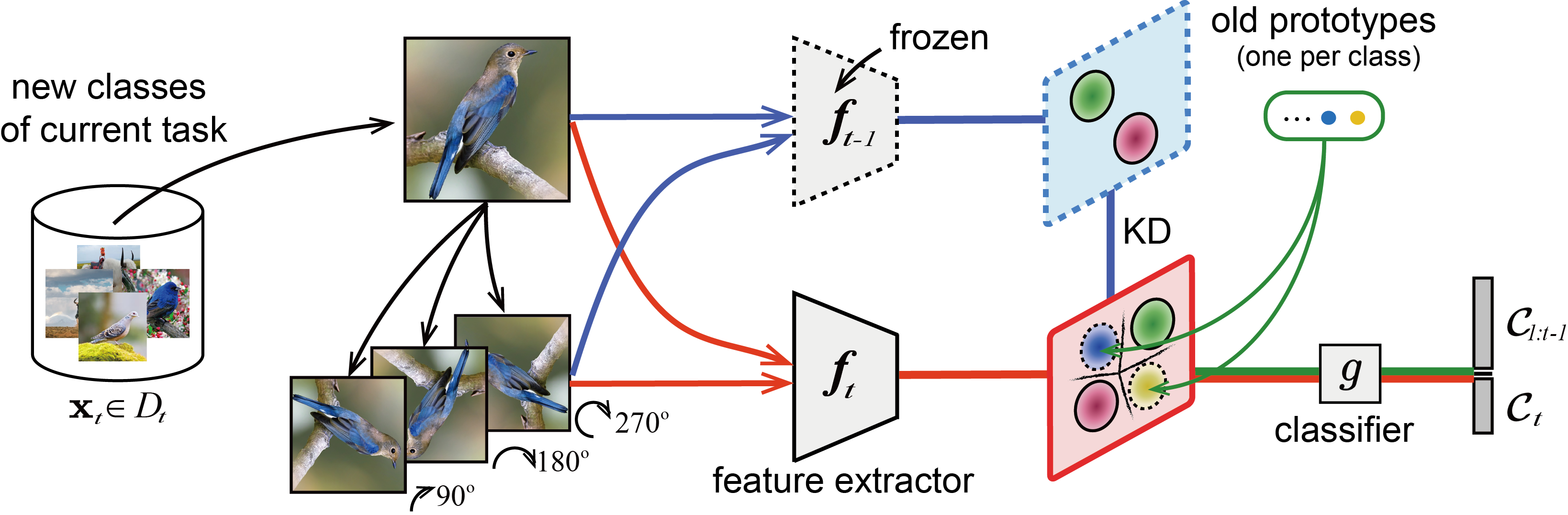}}
		\caption{Illustration of the proposed dual bias reduction framework. Classes of the current task are augmented by rotation based transformation. In the deep feature space, we augment the memorized prototypes explicitly (directly generate augmented feature instances) or implicitly (transforming the original sampling process to a regularization term). A hardness-aware (informative) protoAug strategy is further proposed to compensate for the original protoAug.}
		\label{figure-2}
	\end{center}
	\vskip -0.15in
\end{figure*}

\subsection{Non-Exemplar Class-Incremental Learning}
\textbf{Regularization} based methods can be further divided into two categories, \emph{i.e.,} explicit and implicit regularization, based on what kind of knowledge they aim to preserve. Explicit regularization methods \cite{Kirkpatrick2017OvercomingCF, Zenke2017ContinualLT} identify and penalize the changes of important parameters of the original network when learning new classes. Another line of works \cite{zeng2019continual, lin2022towards, wang2021training, kong2022balancing, lintrgp} avoid interfering with previously learned knowledge by projecting and updating the parameters in the null space of the previous tasks. Yu et al., \cite{Yu2020SemanticDC} found that metric learning based embedding network suffers less forgetting for CIL than softmax-based networks. Rather than directly constraining the parameters of network, implicit regularization approaches \cite{Li2018LearningWF, Kirkpatrick2017OvercomingCF, DharSPWC19, zhu2021calibration, magistrielastic} focus on keeping the input-output behavior of the network based on current training data via knowledge distillation. 

\vspace*{3pt}
\noindent\textbf{Pre-trained model} based methods \cite{wang2022learning, wang2022dualprompt, smith2022coda, villa2022pivot, zhou2023revisiting, liu2023class, cao2024generative} utilize pre-trained frozen models and learn a set of learnable parameters that dynamically instruct models to solve tasks sequentially. L2P \cite{wang2022learning} formulates the problem of learning new tasks as training small prompt parameters attached to a pre-trained frozen model. CODA-Prompt \cite{smith2022coda} adopts a attention-based end-to-end key-query scheme that learns a set of input-conditioned prompts. However, prompting based methods are largely based on pre-trained models, and the information leak both in terms of features and class labels can seriously affect the results of those pre-trained model based methods \cite{kim2023learnability}.
Recently, Kim \emph{et al.}, \cite{kim2023learnability, kim2022multi} and Lin \emph{et al.}, \cite{lin2023class} decompose CIL into out-of-distribution detection and within task prediction.

\subsection{Other studies of Class-Incremental Learning}
There have been some empirical studies \cite{mirzadeh2020understanding, de2022continual, mirzadeh2022wide, kimtheoretical, zhuacta} on incremental learning. Verwimp \emph{et al.}, \cite{verwimp2021rehearsal} revealed the limits of exemplar-based methods. Kim \emph{et al.} \cite{kim2023learnability} studied the learnability of CIL and explored the connection \cite{lee2021continual} between CIL and out-of-distribution \cite{cheng2023average}.  
Besides, some works focus on new CIL settings such as long-tailed \cite{liu2022long}, forgettable \cite{zhao2024continual}, self-supervised \cite{fini2022self, liu2024branch}, federated \cite{guo2024federated} and few-shot \cite{liu2023learnable, yang2022dynamic} CIL. For example, DSN \cite{yang2022dynamic} freezes the backbone and tentatively
expands network nodes to enlarge feature representation capacity for incremental classes.

\section{Problem Statement and Analysis}
A CIL problem involves the sequential learning of tasks that consist of disjoint class sets, and the model has to learn a unified classifier that can classify all seen classes. Formally, at incremental step $t$, a dataset $\mathcal{D}^{t} = \{\bm{x}_{i}^{t}, y_{i}^{t}\}^{n_{t}}_{i=1}$ is given, where $\bm{x}$ is a sample in the input space $\mathcal{X}$ and $y \in \mathcal{C}_{t}$ is its corresponding label. $\mathcal{C}_{t}$ is the class set of task $t$ and the class sets of different tasks are disjoint, \emph{i.e.} $\mathcal{C}_{i} \bigcap \mathcal{C}_{j} = \emptyset$ if $i \neq j$.
To facilitate analysis, we represent the DNN based model with two components: a feature extractor and a unified classifier. Specifically, the feature extractor $f_{\bm{\theta}}: \mathcal{X} \rightarrow \mathcal{Z}$, 
parameterized by $\bm{\theta}$, maps the input $\bm{x}$ into a feature vector $\bm{z} = f_{\bm{\theta}}(\bm{x}) \in \mathbb{R}^{d}$ in the feature space $\mathcal{Z}$; the classifier $g_{\bm{\varphi}}: \mathcal{Z} \rightarrow \mathbb{R}^{|\mathcal{C}_{1:t}|}$, parameterized by $\bm{\varphi}$, produces a probability distribution $g_{\bm{\varphi}}(\bm{z})$ as the prediction for $\bm{x}$.

At incremental step $t$, the general objective is to minimize a predefined loss function $\ell$ (\emph{e.g.}, cross-entropy loss) on new dataset $\mathcal{D}^{t}$ without interfering and with possibly improving on those that were learned previously \cite{Aljundi2019ContinualLI}: 
\begin{equation}\label{eq1}
\begin{aligned}
\min\limits_{\theta,\varphi,\epsilon} \mathbb{E}_{(x,y) \backsim \mathcal{D}^{t}} [&\ell(g_{\varphi}(f_{\theta}(x)), y)] + \sum\nolimits \epsilon_{i}  \\ 
\text{s.t.}~~ \mathbb{E}_{(x,y) \backsim \mathcal{D}^{i}}[ \ell(g_{\varphi}(f_{\theta}(x)), y) -&\ell(g_{\varphi^{t-1}}(f_{\theta^{t-1}}(x)), y) ] \leqslant \epsilon_{i}, \\ \epsilon_{i} \geqslant 0; \forall i &\in \{1,2,...,t-1\}.
\end{aligned}
\end{equation}
The last term $\epsilon = \{\epsilon_{i}\}$ is a slack variable that tolerates a small increase in the loss on datasets of old tasks.

The central challenge of CIL is that data of old classes are assumed to be unavailable, which means that the best configuration of the model for all seen classes must be sought by minimizing $\mathcal{L}_{t}$ on current data $\mathcal{D}^{t}$:
\begin{equation}\label{eq2}
\begin{aligned}
\mathcal{L}_{t} \triangleq \mathbb{E}_{(\bm{x},y) \backsim \mathcal{D}^{t}}[\ell(g_{\bm{\varphi}}(f_{\bm{\theta}}(\bm{x})), y)].
\end{aligned}
\end{equation}
However, as demonstrated in Sec. \ref{sec:introduction}, this would lead to representation bias and classifier bias. On the one hand, it is difficult to learn generic representations which can generalize to classes of both previous and future tasks. As a result, the model updated on $\mathcal{D}^{t}$ would not only suffer from forgetting representations learned on old tasks but also be a bad initialization for future tasks. On the other hand, the decision boundary learned previously can be dramatically distorted, leading to a severely biased classifier. Therefore, addressing the dual bias problem is crucial for CIL.

\section{The Dual Bias Reduction Framework} \label{sec:method}
\textbf{Overview of Framework.} The framework of our method is shown in Fig.~\ref{figure-2}. For representation learning, we leverage rotation based self-supervised transformation to learn diverse and transferable representations for classes in different incremental stages. For classifier learning, we memorize a class-representative prototype for each old class in the deep feature space. When learning a new task, each old prototype is augmented explicitly (directly generating augmented feature instances) or implicitly (transforming the original sampling process to a regularization term) and fed to the unified classifier. In addition, we also use knowledge distillation technique \cite{hinton2015distilling, Hou2019LearningAU}. Note that the conference version \textbf{PASS} (\textbf{P}rototype \textbf{A}ugmentation and \textbf{S}elf-\textbf{S}upervision) \cite{Zhu_2021_CVPR} consists of the original protoAug (Sec. \ref{sec:EPA}) and SST (Sec. \ref{sec:labelAug}), and \textbf{PASS++} denotes the current version which further integrates the newly proposed hardness-aware protoAug (Sec. \ref{sec:haprotoAug}) and multi-view ensemble (Sec. \ref{sec:mve}).

\subsection{Reduce Representation Bias with Self-Supervised Transformation} \label{sec:labelAug}
As we focus on non-exemplar CIL, we intentionally avoid storing old data. To maintain generalization of the learned representations for old classes, existing methods typically restrain the feature extractor from changing \cite{Kirkpatrick2017OvercomingCF, Zenke2017ContinualLT, aljundi2018memory, Li2018LearningWF}. However, this would lead to a trade-off between plasticity and stability \cite{parisi2019continual}, and it would be hard to perform long-step incremental learning. Our high-level idea is to prepare the close-set training for other (previous and future) classes at each incremental stage. To this end, we propose to learn diverse and transferable representations that can be re-used for future tasks and remain unforgotten for task-level generalization at each incremental stage. With diverse and transferable representations, classes from different tasks would be more separated in deep feature space, and it would be easier to find a model that performs well on all tasks.

Technically, inspired by \cite{Gidaris2018UnsupervisedRL, lee20_sla}, we simply learn a unified model by augmenting the current class based on self-supervised transformation (SST). Specifically, for each class, we rotate its training data 90, 180, and 270 degrees to generate 3 additional novel classes, extending the original $\bm{k}$-class problem to a $\bm{4k}$-class problem:
\begin{equation}
\label{eq3}
\bm{x'} = \text{rotate}(\bm{x}, \delta), \delta \in \{0, 90, 180, 270\},\\
\end{equation}
and the augmented sample is assigned a new label $y'$. We denote the new dataset after the above SST at incremental step $t$ as $\mathcal{D}^{t}_{aug} = \{(\bm{x'}_i, y'_i)\}^{n'_t}_{i=1}$. 
Intuitively, SST introduces fine-grained multi-view information into training process, which is beneficial for learning generic and transferable representations. Besides, it only increases the number of labels, thus the number of additional parameters is negligible compared to that of the original network. In Sec. \ref{sec:mve}, we will present a multi-view ensemble strategy which is largely based on the above SST.

\subsection{Reduce Classifier Bias with Prototype Augmentation} \label{sec:protoAug}
As shown in Fig.~\ref{figure-1} and demonstrated in Sec. \ref{sec:introduction}, classifier bias is another problem in CIL. To maintain the decision boundary among old classes, we propose prototype augmentation to restrain the previously learned decision boundary. Concretely, after learning each task, we compute and memorize one prototype (class mean) for each class:
\begin{equation}
\label{eq4}
\begin{aligned}
\bm{\mu}_{k} = \frac{1}{n_{k}}\sum\nolimits_{j=1}^{n_{k}}f_{\bm{\theta}}(\bm{x}_{j}),
\end{aligned}
\end{equation}
where $n_{k}$ is the number of training samples in class $k$.
Then, when learning new classes, the prototype of each learned old class is augmented to generate pseudo-feature instances of old classes. Specifically, both \textit{explicit} and \textit{implicit} augmentation strategies are designed in this work as follows.

\subsubsection{Explicit Prototype Augmentation} \label{sec:EPA}
For each old class $k \in \mathcal{C}_{old}$, pseudo feature instances are generated explicitly by augmenting the corresponding prototype:
\begin{equation}
\label{eq5}
\begin{aligned}
\widetilde{\bm{z}}_{k} = \bm{\mu}_{k} + r \cdot \bm{e},
\end{aligned}
\end{equation}
where $\bm{e} \sim \mathcal{N}(\bm{0}, 1)$ is the derived Gaussian noise which has the same dimension as the prototype. $r$ is a scale to control the uncertainty of augmented prototypes. In particular, the scale $r$ can be pre-defined, or computed as the average variance of the class representations:
\begin{equation}
\label{eq6}
\begin{aligned}
r^2_{t} = \frac{1}{|\mathcal{C}_{old}|+|\mathcal{C}_{t}|}(|\mathcal{C}_{old}| \cdot r^2_{t-1} + \frac{1}{d} \sum\nolimits_{k=1}^{\mathcal{C}_{new}}\textit{tr}(\bm{\Sigma}_{k})),
\end{aligned}
\end{equation}
where $|\mathcal{C}_{old}|$ and $\mathcal{C}_{new}$ represent the number of old classes and new classes at stage $t$, respectively. $d$ is the dimensionality of the deep feature space. $\bm{\Sigma}_{k}$ is the covariance matrix of the features from class $k$ at stage $t$, and the \textit{tr} operation computes the trace of a matrix. We observed that the $r_{t}$ changes slightly at different stages in the course of a CIL experiment. Therefore, we only compute and use the average variance of the features in the first task: 
\begin{equation}
\label{eq7}
\begin{aligned}
r^2 = r^2_{1} =  \frac{1}{|\mathcal{C}_{t=1}| \cdot d}\sum\nolimits_{k=1}^{|\mathcal{C}_{t=1}|}\textit{tr}(\bm{\Sigma}_{k}). 
\end{aligned}
\end{equation}
Finally, the feature instances of new classes and pseudo feature instances of old classes are fed to the unified classifier to maintain the discrimination and balance between old and new classes. More specifically, suppose $M$ deep feature instances are generated for each old class and cross-entropy loss is used, the learning objective can be detailed as:
\begin{equation}
	\begin{aligned}
	\label{eq8}
	\mathcal{L}_{t} &= \underbrace{ \frac{1}{n'_{t}} \sum_{i=1}^{n'_{t}} -\text{log} \left( \frac{e^{\bm{\varphi}_{y_{i}}^\top \bm{z}_{i} + b_{y_{i}}}}{\sum_{c=1}^{|\mathcal{C}_{all}|} e^{\bm{\varphi}_{c}^\top \bm{z}_{i} + b_{c}}} \right) }_{\mathcal{L}_{t, new}:~\text{loss on real features of new classes}} \\ &+ \underbrace{\frac{1}{|\mathcal{C}_{old}|} \sum_{k=1}^{|\mathcal{C}_{old}|} \frac{1}{M} \sum_{m=1}^{M} -\text{log} \left(\frac{e^{\bm{\varphi}_{k}^\top \widetilde{\bm{z}}_{k,m} + b_{k}}}{\sum_{c=1}^{|\mathcal{C}_{all}|} e^{\bm{\varphi}_{c}^\top \widetilde{\bm{z}}_{k,m} + b_{c}}} \right) }_{\mathcal{L}_{t, old}:~\text{loss on generated features of old classes}},
	\end{aligned}
\end{equation}
where $\widetilde{\bm{z}}_{k}$ represents the feature instances augmented for old class $k \in \mathcal{C}_{old}$, $n'_{t}$ is the number of training samples of the current task, and $|\mathcal{C}_{all}| = |\mathcal{C}_{old}| + |\mathcal{C}_{t}|$ is the number of all seen classes at stage $t$. $\bm{\varphi} = [\bm{\varphi}_{1},...,\bm{\varphi}_{|\mathcal{C}_{all}|}]^\top \in \mathcal{R}^{|\mathcal{C}_{all}| \times d}$ and $b = [b_{1},...,b_{|\mathcal{C}_{all}|}]^\top \in \mathcal{R}^{|\mathcal{C}_{all}|}$ are the weight matrix and bias vector of the last fully connected layer, respectively.

\subsubsection{Implicit Prototype Augmentation} \label{sec:IPA}
In CIL, the second term in Eq.~(\ref{eq8}), $\mathcal{L}_{t, old}$, might be computationally inefficient when $M$ and $|\mathcal{C}_{old}|$ are large. Inspired by \cite{maaten2013learning, wang2021regularizing}, we consider the case that $M$ grows to infinity and find an easy-to-compute way to implicitly generate infinite feature instances for old classes, leading to more efficient
implementation.

\vspace*{5pt}
\noindent
\textbf{Upper bound of $\mathcal{L}_{t, old}$.} Formally, for each old class $k \in \mathcal{C}_{old}$, we can generate $M$ instances in the deep feature space from its distribution, \emph{i.e.}, $\widetilde{\bm{z}}_{k} \backsim \mathcal{N}(\bm{\mu}_{k}, \gamma\bm{\Sigma}_{k})$, in which $\gamma$ is a non-negative coefficient. 
In the case of $M \to \infty$, the second term in Eq.~(\ref{eq8}) can be derived as below:
\begin{equation}
\begin{aligned}
	\label{eq9}
	\mathcal{L}_{t, old} &= \frac{1}{|\mathcal{C}_{old}|} \sum_{k=1}^{|\mathcal{C}_{old}|} \mathbb{E}_{\widetilde{\bm{z}}_{k}} \left[ -\text{log} \left( \frac{e^{\bm{\varphi}_{k}^\top \widetilde{\bm{z}}_{k} + b_{k}}}{\sum_{c=1}^{|\mathcal{C}_{all}|} e^{\bm{\varphi}_{c}^\top \widetilde{\bm{z}}_{k} + b_{c}}} \right) \right] \\ &= \frac{1}{|\mathcal{C}_{old}|} \sum_{k=1}^{|\mathcal{C}_{old}|} \mathbb{E}_{\widetilde{\bm{z}}_{k}} \left[ \text{log} \left( \sum_{c=1}^{|\mathcal{C}_{all}|} e^{(\bm{\varphi}_{c}^\top-\bm{\varphi}_{k}^\top) \widetilde{\bm{z}}_{k} + (b_{c} - b_{k})} \right) \right] \\ &\leqslant \frac{1}{|\mathcal{C}_{old}|} \sum_{k=1}^{|\mathcal{C}_{old}|} \text{log} \left( \mathbb{E}_{\widetilde{\bm{z}}_{k}} \left[ \sum_{c=1}^{|\mathcal{C}_{all}|} e^{(\bm{\varphi}_{c}^\top-\bm{\varphi}_{k}^\top) \widetilde{\bm{z}}_{k} + (b_{c} - b_{k})} \right] \right) \\ &= \frac{1}{|\mathcal{C}_{old}|} \sum_{k=1}^{|\mathcal{C}_{old}|} \text{log} \left( \sum_{c=1}^{|\mathcal{C}_{all}|} e^{\bm{v}_{c,k}^\top \bm{\mu}_{k} + (b_{c} - b_{k}) + \frac{\gamma}{2} \bm{v}_{c,k}^\top \bm{\Sigma}_{k} v_{c,k} } \right).
\end{aligned}
\end{equation}
In above equation, $\bm{v}_{c,k} = \bm{\varphi}_{c}-\bm{\varphi}_{k}$. The inequality is based on Jensen's inequality $\mathbb{E}[\text{log}(X)] \leqslant \text{log}\mathbb{E}[X]$, and the last equality is obtained by using the moment-generating function $\mathbb{E}[e^{tX}] = e^{t\mu+\frac{1}{2}\sigma^{2}t^{2}}$, $X \backsim \mathcal{N}(\mu, \sigma^{2})$. 
As can be seen, Eq.~(\ref{eq9}) is an upper bound of original $\mathcal{L}_{t, old}$, which provides an elegant and much more efficient way to \textbf{implicitly} (without directly generating augmented feature instances) generate \textbf{infinite} instances in the deep feature space for old classes. The $\mathcal{L}_{t, old}$ in Eq.~(\ref{eq8}) can be write in the cross-entropy loss form:
\begin{equation}
\begin{aligned}
\label{eq10}
\mathcal{L}_{t, old} = \frac{1}{|\mathcal{C}_{old}|} \sum_{k=1}^{|\mathcal{C}_{old}|} -\text{log} \left( \frac{e^{\bm{\varphi}_{k}^\top \bm{\mu}_{k} + b_{k}}}{\sum_{c=1}^{|\mathcal{C}_{all}|} e^{\bm{\varphi}_{c}^\top \bm{\mu}_{k} + b_{c} + \frac{\gamma}{2}\bm{v}_{c,k}^\top \bm{\Sigma}_{k} v_{c,k}}} \right)
\end{aligned}.
\end{equation}
Intuitively, $\mathcal{L}_{t, old}$ implicitly performs augmentation for $\bm{\mu}_{k}$ based on $\bm{\Sigma}_{k}$. However, we do not need to save the original covariance matrix for each class. Actually, as shown in our experiments, using stored radius of the Gaussian distribution under the identity covariance matrix assumption for implicit augmentation also performs well.

\subsubsection{Hardness-Aware Prototype Augmentation} \label{sec:haprotoAug}
The protoAug generates feature instances of old classes by augmenting the prototypes based on the average variance $r$. However, there are still some shortcomings in the way protoAug maintains the decision boundary:
\begin{itemize}
	\item Despite that Gaussian is commonly used to approximate feature distribution, it may not be able to accurately characterize the geometry of class-wise feature distribution.
	\item Based on Gaussian sampling, the generated feature instances of old classes are mostly around the prototype, therefore some of them may lack direct information for maintaining decision boundary.
\end{itemize}
\begin{figure}[t]
	\begin{center}
		\centerline{\includegraphics[width=\columnwidth]{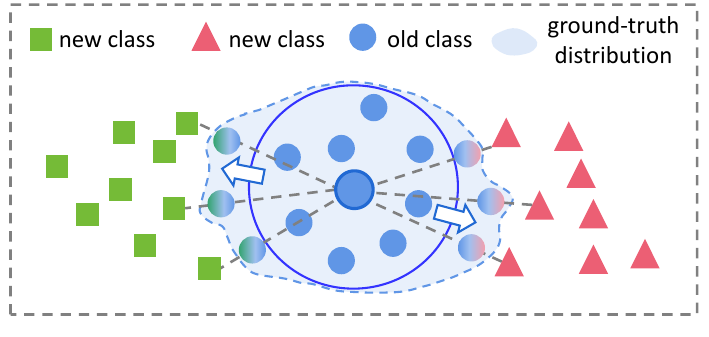}}
		\vskip -0.1in
		\caption{Illustration of hardness-aware prototype augmentation. The constructed hard feature instances provide compensable information to refine the estimated distribution.}
		\label{figure-3}
	\end{center}
	\vskip -0.2in
\end{figure} 
Those two issues limit the effectiveness of protoAug for maintaining previously learned knowledge in CIL.
In light of this, we further propose an adaptive hardness-aware protoAug strategy, which generates informative and hard feature instances near the decision boundary to enhance and compensate for the original protoAug, as illustrated in Fig.~\ref{figure-3}.

Formally, for each learned old class $k \in \mathcal{C}_{old}$, we sort the cosine distance $d(\cdot)$ between feature instances of new classes $\{\bm{z}_{i}\}^{n_{t}}_{i=1}$ and old prototype $\bm{\mu}_{k}$, and get the nearest new sample $\bm{z}^{*}_{\text{new}} = \argmin\limits_{\bm{z}_{i},i=1,...,n_{t}} \{d(\bm{\mu}_{k}, \bm{z}_{1}),...,d(\bm{\mu}_{k}, \bm{z}_{n_{t}})\}$. Then, hardness-aware feature instance can be generated via mixing prototype and the nearest new instance in each minibatch:
\begin{equation}
\label{eq111}
\begin{aligned}
\widetilde{\bm{z}}_{k,\text{hard}} = \lambda \cdot \bm{\mu}_{k} + (1-\lambda) \cdot \bm{z}^{*}_{\text{new}},
\end{aligned}
\end{equation}
where $\lambda$ is a hyper-parameter which is simply set to be 0.7 in our experiments. A reduction in the distance between prototype-sample pairs will create a rise in the hard level and vice versa. Finally, $\widetilde{\bm{z}}_{k,\text{hard}}$ is assigned the same label as $\bm{\mu}_{k}$ and fed to the classifier along with $\widetilde{\bm{z}}_{k}$ generated by Eq.~(\ref{eq5}). The final feature instance set for old class $k \in \mathcal{C}_{old}$ is $\widetilde{\bm{z}}_{k} \cup \widetilde{\bm{z}}_{k,\text{hard}}$. Similar to the original protoAug, we apply hardness-aware protoAug in minibatch (Algorithm 1), which is simple and easy to implement.
By learning those hard feature instances, the decision boundary can be better maintained.

\subsection{Overall Learning Objective} \label{sec:objective}
With SST for representation bias and protoAug for classifier bias, Fig.~\ref{figure-2} describes the learning process of the dual bias reduction framework. We also used knowledge distillation (KD) \cite{hinton2015distilling, Hou2019LearningAU} for two reasons. First, SST and KD are complementary and focus on different aspects of representation learning. Second, KD can reduce the change of feature extractor, which is crucial for protoAug because it implicitly generates features from old distribution. The total learning objective at each stage $t$ is as follows:
\begin{equation}
	\label{eq11}
	\mathcal{L}_{t} = \mathcal{L}_{t, new} + \alpha \cdot \mathcal{L}_{t, old} + \beta \cdot \mathcal{L}_{t, kd},
\end{equation}
where $\alpha$ and $\beta$ are two hyper-parameters. $\mathcal{L}_{t, new}$ and $\mathcal{L}_{t, old}$ are shown in Eq.~(\ref{eq8}) and Eq.~(\ref{eq10}), respectively. $\mathcal{L}_{t, kd} = \frac{1}{n'_{t}} \sum_{i=1}^{n'_{t}} \Arrowvert f_{\bm{\theta}_{t-1}}(x'_{i}) - f_{\bm{\theta}_{t}}(x'_{i}) \Arrowvert$.
In summary, Algorithm 1 outlines how
the proposed PASS++ trains a given neural network for each incremental task. Note that the protoAug in Algorithm 1 is the explicit one which is based on the $\mathcal{L}_{t, old}$ in Eq.~(\ref{eq8}). Alternatively, one can perform implicit protoAug by using the $\mathcal{L}_{t, old}$ in Eq.~(\ref{eq10}).

\begin{algorithm}[t]
	\caption{Dual bias reduction algorithm for CIL}
	$\bm{\Theta}^{0} = \{ \bm{\theta}^{0}, \bm{\varphi}^{0}\} \leftarrow$ a randomly initialized DNN\;
	$\mathcal{P}^{0} = \emptyset$, $r = 0$\;
	\ForEach{incremental stage $t$}{
		\KwIn{model $\bm{\Theta}^{t-1}$, new data $\mathcal{D}^{t} = \{(\bm{x}_{i}^{t}, y_{i}^{t})\}^{n_{t}}_{i=1}$;}
		\KwOut{model $\bm{\Theta}^{t}$;}
		$\bm{\Theta}^{t} \leftarrow \bm{\Theta}^{t-1}$\;
		\textcolor{gray}{\# perform rotation based SST in input space}
		$\mathcal{D}^{t}_{aug} = \{(\bm{x'}_i, y'_i)\}^{n'_t}_{i=1}$ by Eq.~(\ref{eq3})\;
		\eIf{$t=1$}{
			train $\bm{\Theta}^{t}$ by minimizing $\mathcal{L}^{t}(g_{\bm{\varphi}}(f_{\bm{\theta}}(\bm{x'})), y')$\;
			$p \leftarrow$ compute class mean for classes in $\mathcal{D}^{t}_{aug}$\;
			$r \leftarrow$ compute radial scale using Eq.~(\ref{eq7})\;
		}{
			\textcolor{gray}{\# perform protoAug in deep feature space}
			$\widetilde{\bm{z}}_{k} \cup \widetilde{\bm{z}}_{k,\text{hard}}  \leftarrow$ protoAug for old classes by Eq.~(\ref{eq5}) and Eq.~(\ref{eq111})\;
			train $\bm{\Theta}^{t}$ by minimizing Eq.~(\ref{eq11})\;
			$p \leftarrow$ compute class mean for classes in $\mathcal{D}^{t}_{aug}$\;
		}
		$\mathcal{P}^{t} \leftarrow \mathcal{P}^{t-1} \cup p$\;
		Prediction ensemble for inference using Eq.~(\ref{eq13})\;
	}
\end{algorithm}

\subsection{Multi-view Ensemble for Inference}  \label{sec:mve}
As described in Sec. \ref{sec:labelAug}, we augment the current class via rotation based SST, extending the original $\bm{k}$ classes to $\bm{4k}$ classes. However, the additional $\bm{3k}$ classes are only used to learn better representation, not yet involved in classification at inference time. Actually, those classes provide useful information about different views of a given input. Therefore, we propose multi-view ensemble to leverage them properly. Specifically, at inference time, a given test sample $x$ is rotated to get three other views following Eq.~(\ref{eq3}). Then, the logits output on the corresponding $\bm{k}$ classes of each view are aggregated for the final prediction as follows: 
\begin{equation}\label{eq13}
	P(\bm{x}) = \frac{1}{4} \sum_{\delta} g_{\bm{\varphi}, \delta}(f_{\bm{\theta}}(\bm{x}_{\delta})), \delta \in \{0, 90, 180, 270\},
\end{equation}
where $g_{\bm{\varphi}, \delta}$ denotes classifier weights of the corresponding $\bm{k}$ classes for each view $\delta$. Then $\hat{y} =: \arg\max_{y \in \mathcal{C}_{t}} P(\bm{x})$ can be returned as the predicted class.
With ensemble, 
multi-view information of a given sample is aggregated for more robust prediction, yielding remarkably better accuracy. In this strategy, the backbone is unified and only the classification weights are involved in the ensemble, which is lightweight and different from the traditional model ensemble \cite{ganaie2022ensemble}. Besides, our strategy is also different from the vanilla ensemble, which feeds each view of an image into the classifier $g_{\bm{\varphi}}$ with the same $k$ class nodes. More discussion and experimental comparison can be found in Sec. \ref{sec:analysis}.

\begin{figure*}[t]
	\begin{center}
		\centerline{\includegraphics[width=0.999\textwidth]{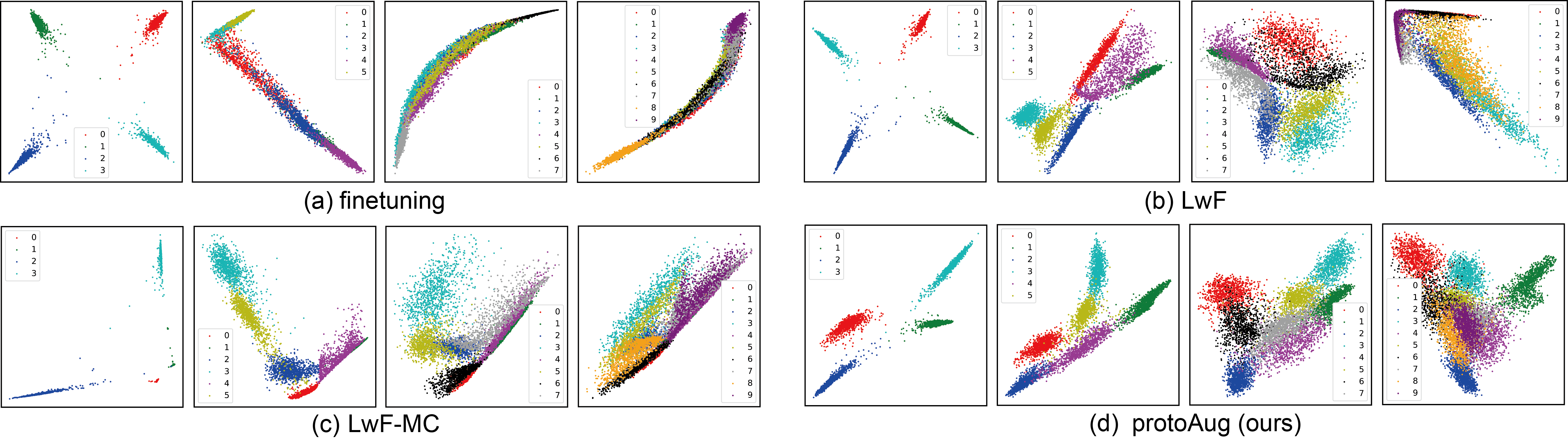}}
		\vskip -0.05in
		\caption{TSNE \cite{van2008visualizing} visualization of class representations in the feature space when learning MNIST \cite{LeCun2005TheMD} incrementally. The outputted features are \textit{2-dimensional} which is suitable for visualization. Best viewed in color.}
		\label{figure-4}
	\end{center}
	\vskip -0.2in
\end{figure*}
\subsection{Integrate with Pre-trained Model} \label{sec:pretrained}
Recently, some works use pre-trained models for CIL \cite{wang2022learning, lin2023class, kim2023learnability, kim2022multi}. Here we discuss how the proposed method can be easily integrated with pre-trained models like ViT \cite{dosovitskiy2020image}. To leverage the pre-trained model while adapting to new knowledge, we insert a low-rank adapter (LoRA) \cite{hulora} at each transformer layer. Specifically, LoRA composes of two rank decomposition matrices $\mathbf{B} \in \mathbb{R}^{u \times r}$ and $\mathbf{A} \in \mathbb{R}^{r \times v}$ where $r \in \mathbb{N}$ is the rank and $r \ll {\rm min}(u, v)$. $v$ and $u$ are the dimensionality of the input $\mathbf{\hat{x}} \in \mathbb{R}^{v}$ for current layer and hidden features, respectively. The modified forward pass with LoRA becomes:
\begin{equation}
	\mathbf{z} = (\mathbf{W} + \mathbf{BA})\mathbf{\hat{x}} = \mathbf{W}\mathbf{\hat{x}} + \mathbf{BA}\mathbf{\hat{x}},
\end{equation}
where $\mathbf{z} \in \mathbb{R}^{u}$ is the output, which will be the input of the next layer after passing non-linear activation. During the training stage, the original parameters $\mathbf{W}$ remain frozen, while only $\mathbf{A}$ and $\mathbf{B}$ are trainable, which is low-cost and parameter efficient. For CIL, we just train the LoRA modules using the overall learning objective described in Sec. \ref{sec:objective}.

\section{Preliminary Experiments} \label{sec:toy-experiment}
In this section, we provide visualization and proof-of-concept experiments to intuitively understand the effectiveness of protoAug and SST for CIL, respectively. 

\vspace*{5pt}
\noindent\textbf{2D Visualization of protoAug.} To provide an illustration of protoAug, we conduct experiments on MNIST \cite{LeCun2005TheMD} with a \textit{2-dimensional} feature space which is suitable for visualization. SST is not applied here since the effect of protoAug is the focus of this experiment. We start from a ResNet-18 model \cite{he2016deep} trained on 4 classes and the remaining 6 classes are continually added in 3 phases. We compare our method with fine-tuning, LwF \cite{Li2018LearningWF}, and LwF-MC (binary cross entropy based) \cite{Rebuffi2017iCaRLIC}. As shown in Fig.~\ref{figure-4}, the distribution of old classes is dramatically changed in fine-tuning, and the previously learned decision boundary suffers from increasingly serious distortion during the incremental learning process, resulting in catastrophic forgetting. LwF \cite{Li2018LearningWF} and LwF-MC \cite{Rebuffi2017iCaRLIC} reduce such distortion to a certain degree. However, there are still obvious overlaps of distribution from different classes, making it difficult to correctly classify them. Contrarily, our method can maintain the distribution of old classes when learning new classes by restraining the decision boundary via protoAug, successfully reducing the forgetting phenomenon.

\vspace*{5pt}
\noindent\textbf{A closer look at SST for CIL.} We train ResNet-18 \cite{he2016deep} for classifying CIFAR-10 and CIFAR-100 \cite{krizhevsky2009learning}. Following the zero-cost CIL paradigm \cite{LeeLLS18, MensinkVPC13}, we first train a classification model on some (4 for CIFAR-10 and 40 for CIFAR-100) base classes. Then a nearest-class-mean (NCM) classifier is built on the pre-trained feature extractor to classify both base and new classes incrementally. We train all the models for 100 epochs with batch size 64 and Adam \cite{kingma2015adam} optimizer with 0.001 initial learning rate. The learning rate is multiplied by 0.1 after 45 and 90 epochs. 
\begin{table}[t]
	\vskip 0.05in
	\caption{Results of zero-cost class incremental learning. The model is tested using nearest-class-mean classifier. Best results are highlighted in bold.}
	\vskip -0.13in
	\label{table-1}
	\begin{center}
		\renewcommand\tabcolsep{2.5pt}
		\newcommand{\tabincell}[2]{\begin{tabular}{@{}#1@{}}#2\end{tabular}}
		\scalebox{1}{
			\renewcommand{\arraystretch}{1.2}
			\begin{tabular}{llcccccc}
				\toprule[1.3pt]	
				Dataset &Method  & Old-All	&Old-Old	&New-All	&New-New	&All\\
				\midrule
				\multirow{2}*{\tabincell{c}{CIFAR-10}}
				&Baseline & 86.75  & 96.20  & 24.86  & 36.43  & 49.62 \\
				&w/ SST & \textbf{89.25}  & \textbf{96.85}  & \textbf{38.40}  & \textbf{44.38}  & \textbf{58.74}\\
				\midrule
				\multirow{2}*{\tabincell{c}{CIFAR-100}}
				&Baseline & \textbf{66.40}  & 76.60  & 31.00  & 40.57  & 45.16\\
				&w/ SST & 65.64  & \textbf{80.55}  & \textbf{46.93}  & \textbf{52.11}  & \textbf{54.42}\\
				\bottomrule[1.3pt]
		\end{tabular}}
	\end{center}
	\vskip -0.1in
\end{table}
\begin{figure}[!t]
	\begin{center}
		\centerline{\includegraphics[width=0.95\columnwidth]{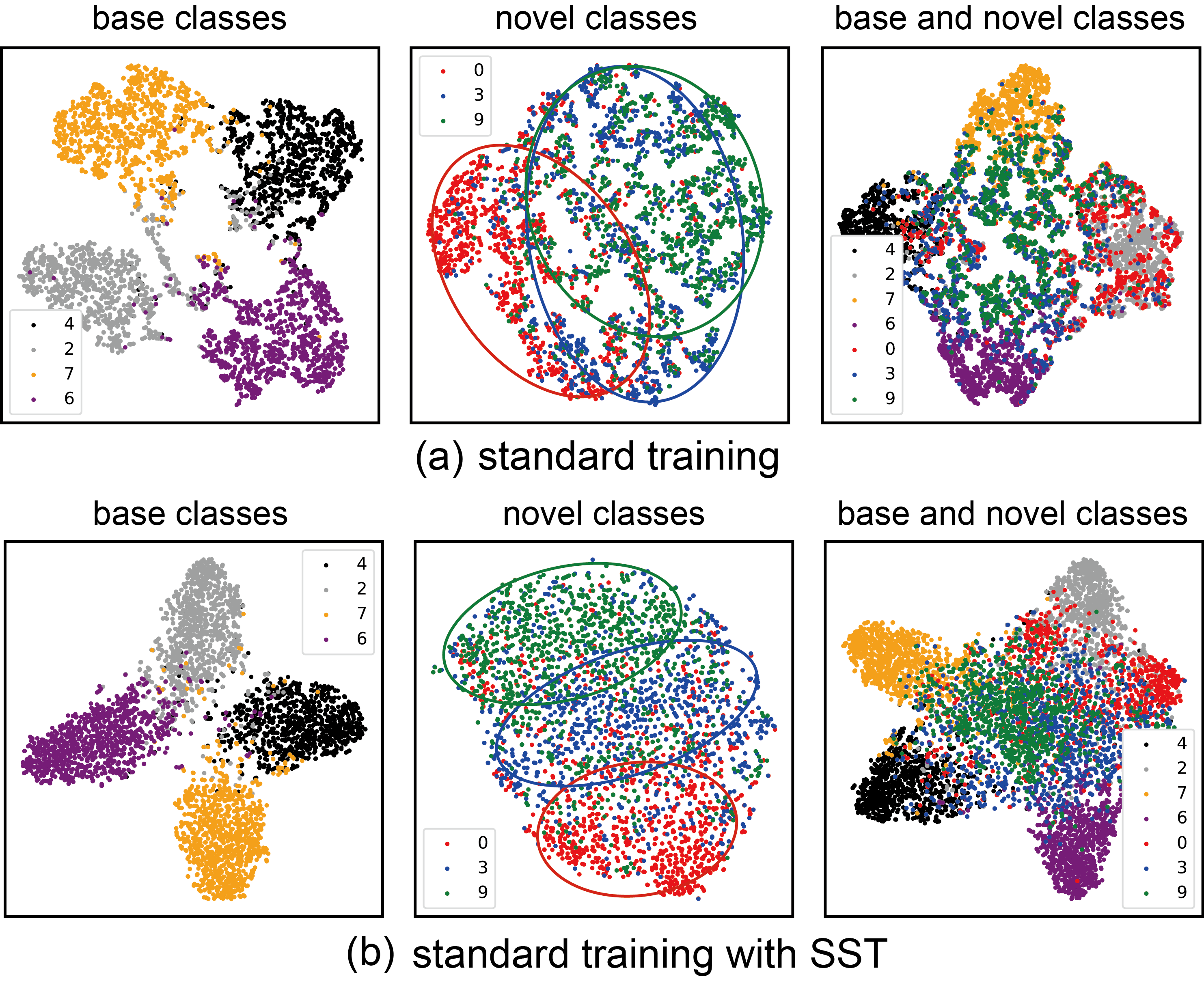}}
		\vskip -0.03in
		\caption{TSNE \cite{van2008visualizing} visualization shows that SST improves the separation of novel classes, reducing the overlap between base and novel classes.}
		\label{figure-5}
	\end{center}
	\vskip -0.35in
\end{figure}
In Table \ref{table-1}, we provide five kinds of accuracies, i.e., 	\textbf{\emph{Old-All}} (average accuracy of old/base classes within all classes),	\textbf{\emph{Old-Old}} (average accuracy of old/base classes within old/base classes),		\textbf{\emph{New-All}}	(average accuracy of new classes within all classes),	\textbf{\emph{New-New}}	(average accuracy of new classes within new classes),	\textbf{\emph{All}} (average accuracy of all classes in the dataset).
We can observe that the performance of old classes is generally maintained after applying SST, while the accuracy of novel classes can be significantly improved with SST, demonstrating the suitability and effectiveness of SST for CIL. An intuitive explanation for the effectiveness of SST on novel class generalization is it improves the separation of novel classes, as shown in Fig.~\ref{figure-5}.

\begin{table*}[t]
	\caption{Comparisons of average and last incremental accuracies (\%). For exemplar-based methods, $R=15$ on CIFAR-100 and TinyImageNet, and $R=10$ on ImageNet-Subset. The baselines are divided into two groups: the first group contains exemplar-based methods, and the second group contains non-exemplar methods. Best results are highlighted in bold.}
	\vskip -0.13in
	\label{table-2}
	\begin{center}
		\renewcommand\tabcolsep{3.5pt}
		\newcommand{\tabincell}[2]{\begin{tabular}{@{}#1@{}}#2\end{tabular}}
		\scalebox{1}{
			\renewcommand{\arraystretch}{1.24}
			\begin{tabular}{lcccccccccccccccccc}
				\toprule[1.3pt]
				\multirow{3}*{\tabincell{c}{\textbf{Method}}} &
				\multicolumn{6}{c}{\makecell{\ \textbf{CIFAR-100}}} & \multicolumn{6}{c}{\makecell{\ \textbf{TinyImageNet}}}& \multicolumn{6}{c}{\makecell{\ \textbf{ImageNet-Subset}}}\\
				\cmidrule(lr){2-7} \cmidrule(lr){8-13} \cmidrule(lr){14-19}		
				&\multicolumn{2}{c}{\makecell{$T = 5$}} & \multicolumn{2}{c}{\makecell{$T = 10$}} & \multicolumn{2}{c}{\makecell{$T = 20$}}  &\multicolumn{2}{c}{\makecell{$T = 5$}} & \multicolumn{2}{c}{\makecell{$T = 10$}} & \multicolumn{2}{c}{\makecell{$T = 20$}}
				&\multicolumn{2}{c}{\makecell{$T = 5$}} & \multicolumn{2}{c}{\makecell{$T = 10$}} & \multicolumn{2}{c}{\makecell{$T = 20$}}  \\
				\cmidrule(lr){2-3} \cmidrule(lr){4-5} \cmidrule(lr){6-7} \cmidrule(lr){8-9} \cmidrule(lr){10-11} \cmidrule(lr){12-13} \cmidrule(lr){14-15} \cmidrule(lr){16-17} \cmidrule(lr){18-19}
				& Avg &Last & Avg &Last & Avg &Last & Avg &Last & Avg &Last & Avg &Last & Avg &Last & Avg &Last & Avg &Last \\
				\midrule
				iCaRL-cnn \cite{Rebuffi2017iCaRLIC}&49.42  &37.24 &43.95  &34.72 &42.70  &35.20  &31.94 &19.07 &25.04 &15.98  &21.87 &16.89  &39.52 &21.92  &31.40  &18.04 &24.00 &14.24 \\
				iCaRL-ncm \cite{Rebuffi2017iCaRLIC}&59.25  &48.11 &53.50  &44.43 &51.69 &41.08  &44.24 &31.94 &38.24 &27.85  &33.63 &25.36  &60.21 &44.14  &52.79  &36.84 &40.33 &28.76 \\
				BiC \cite{Wu2019LargeSI}&61.48  &47.75 &55.20  &41.81 &51.32  &37.69  &46.37 &30.55 &39.02 &24.56  &34.43 &21.18  &63.35 &47.94  &56.60  &35.76 &50.97 &29.54 \\
				UCIR \cite{Hou2019LearningAU}&67.06  &57.02 &64.39  &54.29 &59.07  &47.97  & 52.39 &41.46 &49.42 &38.39  &43.21 &33.14  & 65.23 &50.94  &60.08  &43.96 &48.87 &31.74 \\
				PODnet \cite{Douillard2020SmallTaskIL}&67.08  &55.78 &61.39  &49.45 & 57.73  &47.31  &\textbf{54.32} &\textbf{45.01} &\textbf{52.45}  &\textbf{41.17}  &\textbf{48.08}  & 35.97 &\textbf{70.12}   & 57.08 &\textbf{65.32}   &49.76 &53.99   &36.08 \\
				DER \cite{yan2021dynamically}&\textbf{68.99}  &\textbf{61.11} &\textbf{66.73}  &\textbf{57.38} &\textbf{63.53}  &\textbf{51.64}  &50.69 &43.67 &46.54 &38.65  &43.82 &\textbf{36.17}  &67.29 &\textbf{62.18}  & 63.36 &\textbf{54.74} &\textbf{61.91} &\textbf{53.16} \\
				\midrule
				LwF-MC \cite{Li2018LearningWF}&33.38  &17.39 &26.01  &12.15 &19.70  &9.40  &34.91 &18.86 &21.38 &10.31 &13.68 &5.72 &51.41 &32.98 &35.79 &18.46 &19.61 &9.14  \\
				LwM \cite{DharSPWC19}&39.60  &15.71 &30.24  &8.59  &20.54  &3.70  &37.32 &15.29 &20.47 &7.77  &12.55 &2.82  &48.39 &21.34  &32.57  &10.22 &18.86 &3.84 \\
				CCIL \cite{zhu2021calibration}&60.65  &45.57 &43.58  &24.37 &38.05  &16.91  &36.72 &26.18 &27.64 &17.37  &16.28 &10.93  &53.39 &33.44  &41.11  &22.22 &28.96 &8.63 \\
				\textcolor{blue}{DSN} \cite{yan2021dynamically} &50.43 &36.19 &57.19 &45.80 & 54.26 & 43.16 & 45.26 &33.41 &42.69 & 31.87 &38.93 & 31.04 &66.27 &53.15 &59.37 &42.03 &47.29 &30.58\\
				IL2A \cite{Zhu_2021_NeurIPS}&66.17  &54.98 &58.20  &45.07 &58.01  &50.90  &47.21 &36.58 &44.69 &34.28  &40.00 &33.62 &68.11 &55.72 &62.30 &45.87 &50.95 &32.28 \\
				SSRE \cite{zhu2022self} &65.21 &56.73 & 64.94 &55.10 & 61.36 &51.17 & 50.39 &42.06 & 48.93 &40.72 & 48.17  &38.05 
				&70.15 &60.26 &66.42 &57.83 &60.12 &44.68 \\
				PASS \cite{Zhu_2021_CVPR}&63.84  &55.67 &59.87  &49.03 &58.06  &48.48  &49.53 &41.58 &47.15  &39.28 &41.99 &32.78 &69.12  &56.02  &63.02 &47.68 &51.40 &30.30 \\
				\rowcolor{mygray}
				PASS++ (Ours) &\textbf{69.12}  &\textbf{59.87} &\textbf{66.50}  &\textbf{57.69} &\textbf{64.32}  &\textbf{53.43}  &\textbf{54.13} &\textbf{46.93} &\textbf{53.14} &\textbf{46.66}  &\textbf{49.70} &\textbf{40.53} &\textbf{73.87} &\textbf{63.66}  &\textbf{71.86} &\textbf{60.90} &\textbf{65.79} &\textbf{49.38} \\
				\bottomrule[1.3pt]
		\end{tabular}}
	\end{center}
	\vskip -0.05in
\end{table*}

\begin{figure*}[t]
	\begin{center}
		\centerline{\includegraphics[width=0.97\textwidth]{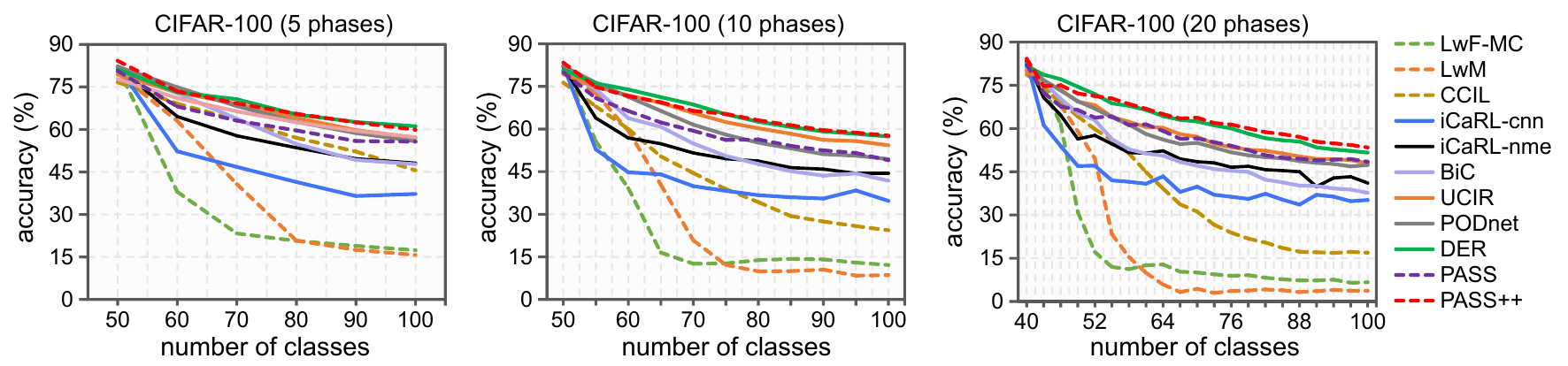}}
		\vskip -0.07in
		\caption{Results of classification accuracy on CIFAR-100, which contains 5, 10 and 20 sequential tasks. Dashed lines represent non-exemplar methods, solid lines denote exemplar-based methods.}
		\label{figure-6}
	\end{center}
	\vskip -0.24in
\end{figure*}
\begin{figure*}[!t]
	\begin{center}
		\centerline{\includegraphics[width=0.97\textwidth]{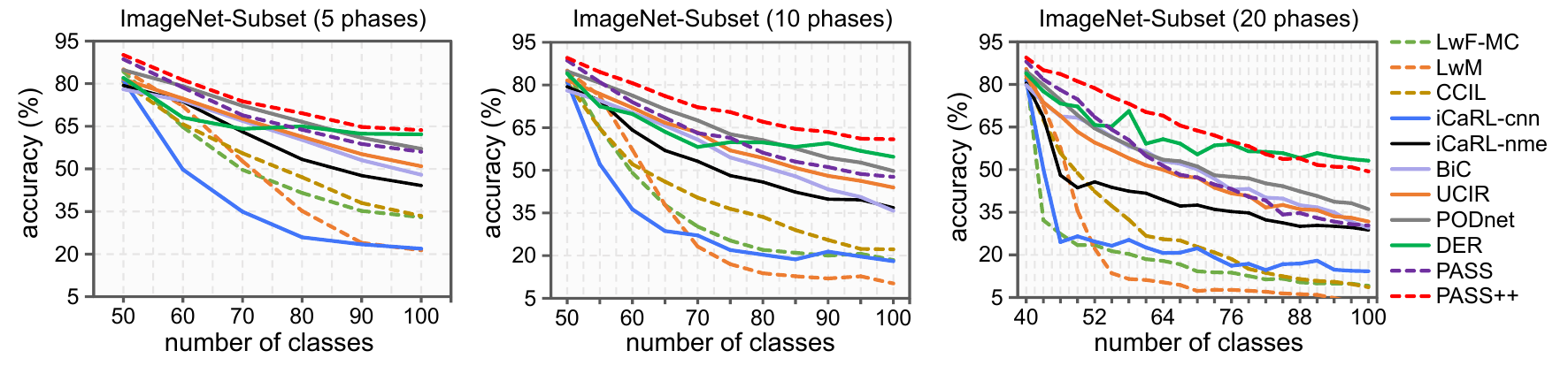}}
		\vskip -0.07in
		\caption{Results of classification accuracy on ImageNet-Subset, which contains 5, 10 and 20 sequential tasks.}
		\label{figure-7}
	\end{center}
	\vskip -0.185in
\end{figure*}

\section{Main Experiments} \label{sec:main-experiment}
\subsection{Setup}
\noindent
\textbf{Dataset and networks.} We perform our experiments on benchmark datasets including CIFAR-100 \cite{krizhevsky2009learning}, TinyImageNet \cite{pouransari2015tiny}, ImageNet-Subset and the
larger and more difficult ImageNet-Full (ImageNet-1k) \cite{deng2009imagenet}. ResNet-18 \cite{he2016deep} is adopted as feature extractor following \cite{yan2021dynamically, boschini2022class, Zhu_2021_CVPR}. 

\vspace*{5pt}
\noindent
\textbf{Comparison methods.} The compared approaches include: \textbf{(1)} \textbf{\emph{Non-exemplar}} methods such as LwF-MC \cite{Li2018LearningWF,Rebuffi2017iCaRLIC}, LwM \cite{DharSPWC19}, CCIL \cite{zhu2021calibration}, NCM \cite{MensinkVPC13}, IL2A \cite{Zhu_2021_NeurIPS}, SSRE \cite{zhu2022self} as well as PASS \cite{Zhu_2021_CVPR}; \textbf{(2)} \textbf{\emph{Exemplar-based}} approaches include iCaRL \cite{Rebuffi2017iCaRLIC}, BiC \cite{Wu2019LargeSI}, UCIR \cite{Hou2019LearningAU}, GeoDL \cite{simon2021learning}, PODnet \cite{Douillard2020SmallTaskIL}, Mnemonics \cite{liu2020mnemonics}, AANets \cite{liu2021adaptive}, CwD \cite{shi2022mimicking}, SSIL \cite{ahn2021ss}, DualNet \cite{pham2021dualnet}, DMIL \cite{tang2022learning}, AFC \cite{kang2022class}, BiMeCo \cite{nie2023bilateral}, EOPC \cite{wen2023optimizing}, DER \cite{yan2021dynamically}, FOSTER \cite{wang2022foster}, Dytox \cite{douillard2022dytox} and DNE \cite{hu2023dense}; \textbf{(3)} \textbf{\emph{Generative exemplar-replay}} methods includes ABD \cite{smith2021always} and R-DFCIL \cite{gao2022r}. Besides, we conduct another set of experiments using a pre-trained DeiT-S/16 \cite{touvron2021training} backbone, and compare with more non-exemplar and exemplar-based method baselines such as OWM \cite{zeng2019continual}, ADAM \cite{zhou2023revisiting}, HATCIL \cite{kimtheoretical}, SLDA \cite{hayes2020lifelong}, L2P \cite{wang2022learning}, A-GEM \cite{chaudhry2018efficient}, EEIL \cite{castro2018end}, GD \cite{lee2019overcoming}, DER++ \cite{buzzega2020dark}, HAL \cite{chaudhry2021using}, 
			FOSTER \cite{wang2022foster}, 
			BEEF \cite{wang2022beef}, 
			MORE \cite{kim2022multi},
			ROW \cite{kim2023learnability} and TPL \cite{lin2023class}.

\vspace*{5pt}
\noindent
\textbf{Evaluation protocol and metrics.} 
The following metrics are used to evaluate the performance: \textbf{(1)} \textbf{\emph{Last accuracy}} $a_{\text{last}}$ is defined as the top-1 accuracy of all learned classes at the final incremental stage in the CIL process. \textbf{(2)} \textbf{\emph{Average accuracy}} is computed as $A_{t}=\frac{1}{t}\sum_{i=1}^{t}a_{i}$, in which $a_{i}$ is the top-1 accuracy of all
the classes that have already been learned at stage $i$. \textbf{(3)} \textbf{\emph{Forgetting}} at step $k$ ($k$ > 1) is calculated as $F_{k}=\frac{1}{k-1}\sum_{i=1}^{k-1}f_{k}^{i}$, where $f_{k}^{i}=\max\limits_{t \in {1,...,k-1}} (a_{t,i}-a_{k,i}),\forall i < k$, and $a_{m,n}$ is the accuracy of task $n$ after training task $m$. 

\vspace*{5pt}
\noindent
\textbf{Implementation details.} We mainly train the model on half of the classes for the first task, and equal classes in the rest phases (denote as $\bm{T}$) following \cite{Hou2019LearningAU, Douillard2020SmallTaskIL, Zhu_2021_CVPR}. Three different incremental settings, \emph{i.e.}, 5, 10 and 20 phases are mainly conducted to evaluate CIL performance in both short and long incremental phases. Besides, to directly compare with results in \cite{yan2021dynamically, gao2022r}, we also report other incremental settings like 2 and 25 phases. For our method, explicit protoAug is used for main experiments. We train all the models for 100 epochs with batch size 64. Adam \cite{kingma2015adam} optimizer is used with 0.001 initial learning rate, which is multiplied by 0.1 after 45 and 90 epochs. The reported results are the average performance of repeating experiments three times.
For exemplar-based approaches, \textit{herd selection} \cite{welling2009herding, Rebuffi2017iCaRLIC} is used to select and store $\bm{R}$ old training samples. In per-trained model based experiments, we set 5 or 10 tasks with an equal number of classes for each dataset. We fine-tune all the models for 10 epochs with batch size 64 and Adam optimizer. The initial learning rate is 0.005, and multiplied by 0.1 after 6 epochs.

\subsection{Comparative Results}\label{sec:main-experiment-explicit}
\noindent\textbf{Comparison with other non-exemplar methods.} In Table~\ref{table-2}, Fig.~\ref{figure-6} and Fig.~\ref{figure-7}, we observe that the accuracy of non-exemplar approaches such as LwF-MC \cite{Rebuffi2017iCaRLIC}, LwM \cite{DharSPWC19} and CCIL \cite{zhu2021calibration} drops drastically along with the increase of incremental phases. This observation aligns with \cite{buzzega2020dark, Hsu2018ReevaluatingCL, Ven2019ThreeSF} and suggests that only constraining old parameters does not suffice to prevent forgetting in CIL. A key limitation is that those methods can not alleviate the overlap between old and new classes in the feature space. Consequently, the two biases in CIL, \emph{i.e.}, representations bias and classifier bias have not been well addressed. Our previous proposal (conference version) PASS \cite{Zhu_2021_CVPR} is a strong baseline that achieves acceptable performance. SSRE \cite{zhu2022self} improves PASS by introducing self-sustaining representation expansion.
However, the limitations described in Sec. \ref{sec:haprotoAug} and Sec. \ref{sec:mve} affect the performance. By addressing those limitations, the proposed enhancements of PASS++ deliver much higher accuracy. Consequently, our method performs much better than other non-exemplar methods in the trend of accuracy curve under different settings.

\vspace*{5pt}
\noindent\textbf{Comparison with exemplar-based methods.} As shown in Table~\ref{table-2}, Fig.~\ref{figure-6}, Fig.~\ref{figure-7}, our method can outperform representative exemplar-based methods such as iCaRL \cite{Rebuffi2017iCaRLIC}, UCIR \cite{Hou2019LearningAU} and PODnet \cite{Douillard2020SmallTaskIL}. Besides, we also make comparison with some recent methods under the setting of $R=20$ in Table~\ref{table-3}, where the results are mostly from their original papers. We can observe that our method performs comparably with those exemplar-based methods. 
The above results are invigorating because existing works \cite{buzzega2020dark} have suggested that data replay, as well as re-balance strategy, are crucial for achieving solid performance in CIL. While our method demonstrates that an class incremental learner can indeed perform well without storing any old examples.

\begin{table}[t]
	\vskip 0.05in
	\caption{Average accuacies (\%) on CIFAR-100 and ImageNet-Subset. Exemplar-based methods save 20/class samples.}
	\vskip -0.13in
	\label{table-3}
	\begin{center}
		\renewcommand\tabcolsep{2.9pt}
		\newcommand{\tabincell}[2]{\begin{tabular}{@{}#1@{}}#2\end{tabular}}
		\scalebox{1}{
			\renewcommand{\arraystretch}{1.2}
			\begin{tabular}{lcccccc}
				\toprule[1.3pt]
				\multirow{2}*{\tabincell{c}{\textbf{Method}}} &	
				\multicolumn{3}{c}{\makecell{CIFAR-100}} & \multicolumn{3}{c}{\makecell{ImageNet-Subset}}\\
				\cmidrule(lr){2-4} \cmidrule(lr){5-7}
				& $T=5$ &$T=10$ & $T=25$ & $T=5$ &$T=10$ & $T=25$\\
				\midrule
				iCaRL \cite{Rebuffi2017iCaRLIC}& 57.12 &52.66 &48.22 &65.44 &59.88 &52.97 \\
				LUCIR \cite{Rebuffi2017iCaRLIC}& 63.17&60.14&57.54&70.84&68.32&61.44\\
				PODNet \cite{Douillard2020SmallTaskIL}&64.83&63.19&60.72&75.54&74.33&68.31\\
				GeoDL \cite{simon2021learning}&65.14 &65.03 &63.12 &73.87 &73.55 &71.72 \\
				Mnemonics \cite{liu2020mnemonics}  &63.34&62.28&60.96&72.58&71.37&69.74\\
				AANets \cite{liu2021adaptive}  &66.31 &64.31 &62.31 &76.96 &75.58 &71.78\\
				CwD \cite{shi2022mimicking} &67.44&64.64&62.24&76.91&74.34&67.42\\
				SSIL \cite{ahn2021ss} & 63.02 &61.52 &58.02 &-- &-- &--\\
				DMIL \cite{tang2022learning}&68.01  &66.47 &--   &{77.20} &{76.76} &--\\
				AFC \cite{kang2022class}&66.49  &64.98 &63.89  &76.87 &75.75 &73.34\\
				DualNet \cite{pham2021dualnet}&68.01 &63.42 &63.22&71.36&67.21&66.35\\
				BiMeCo \cite{nie2023bilateral} &{69.87}&{66.82}&{64.16}&72.87&69.91&67.85\\
				EOPC \cite{wen2023optimizing} &67.55 &65.54 &61.82 &78.95 &74.99 &70.10\\
				DER \cite{yan2021dynamically}&  71.69 &70.42 &-- &76.90 &-- &--\\
				FOSTER \cite{wang2022foster}&  72.90 & 67.95 &63.83 &75.85 &-- &--\\
				Dytox \cite{douillard2022dytox}&  71.55 &--&68.31 &75.54 &-- &--\\
				DNE \cite{hu2023dense}& 74.86 & 74.20 &-- &78.56 &-- &--\\
				\midrule
				\cellcolor{mygray}PASS++ (Ours) &\cellcolor{mygray}69.12  &\cellcolor{mygray}66.50  & \cellcolor{mygray}60.63 &\cellcolor{mygray}73.83 &\cellcolor{mygray}71.86 &\cellcolor{mygray}64.13 \\
				\bottomrule[1.3pt]
		\end{tabular}}
	\end{center}
\end{table}
\begin{table}[t]
	\caption{Comparisons with generative exemplar-based methods. Baselines results are come from \cite{gao2022r}.}
	\vskip -0.13in
	\label{table-4}
	\begin{center}
		\renewcommand\tabcolsep{5.6pt}
		\newcommand{\tabincell}[2]{\begin{tabular}{@{}#1@{}}#2\end{tabular}}
		\scalebox{1}{
			\renewcommand{\arraystretch}{1.2}
			\begin{tabular}{lcccccc}
				\toprule[1.3pt]	
				\multirow{2}*{\tabincell{c}{\textbf{Method}}} &	
				\multicolumn{2}{c}{\makecell{$T = 5$}} & \multicolumn{2}{c}{\makecell{$T = 10$}} & \multicolumn{2}{c}{\makecell{$T = 25$}}\\
				\cmidrule(lr){2-3} \cmidrule(lr){4-5} \cmidrule(lr){6-7}
				& Avg &Last & Avg &Last & Avg &Last\\
				\midrule
				&\multicolumn{6}{c}{\makecell{CIFAR-100}} \\
				\cmidrule(lr){2-7}
				UCIR-DF \cite{Hou2019LearningAU}&57.82 &39.49 &48.69 & 25.54  & 33.33 & 9.62 \\
				PODNet-DF \cite{Douillard2020SmallTaskIL}&56.85  &40.54 &  52.61  &33.57 &43.23 &20.18\\
				ABD \cite{smith2021always}&62.40  &50.55 &58.97   &43.65   &48.91 &25.27\\
				R-DFCIL \cite{gao2022r}&64.78  &54.76 &61.71   &49.70  & 49.95 & 30.01\\
				\rowcolor{mygray}
				PASS++ (Ours) &\textbf{69.12}  &\textbf{59.87} &\textbf{66.50}  &\textbf{57.69} &\textbf{60.63}   &\textbf{48.70} \\
				\midrule
				&\multicolumn{6}{c}{\makecell{TinyImageNet}}\\
				\cmidrule(lr){2-7}
				ABD \cite{smith2021always}&44.55  &33.18 & 41.64   &27.34  & 34.47 & 16.46\\
				R-DFCIL \cite{gao2022r}&{48.91}  &{40.44} &{47.60}   &{38.19}  &{40.85} & {27.29}\\
				\rowcolor{mygray}
				PASS++ (Ours) &\textbf{54.13} &\textbf{46.93} &\textbf{53.14} &\textbf{46.66} &\textbf{47.81}   &\textbf{37.86} \\
				\bottomrule[1.3pt]
		\end{tabular}}
	\end{center}
	\vskip -0.1in
\end{table}
\begin{table}[t]
	\caption{Compare our method with exemplar-based methods on ImageNet-Full under 10 and 25 phases.}
	\vskip -0.13in
	\label{table-5}
	\begin{center}
		\renewcommand\tabcolsep{6.5pt}
		\newcommand{\tabincell}[2]{\begin{tabular}{@{}#1@{}}#2\end{tabular}}
		\scalebox{1}{
			\renewcommand{\arraystretch}{1.2}
			\begin{tabular}{clcccc}
				\toprule[1.3pt]
				\multirow{2}*{\tabincell{c}{\textbf{Exemplar}}} &	
				\multirow{2}*{\tabincell{c}{\textbf{Method}}} &	
				\multicolumn{2}{c}{\makecell{$T = 10$}} & \multicolumn{2}{c}{\makecell{$T = 25$}}\\
				\cmidrule(lr){3-4} \cmidrule(lr){5-6}
				&& Avg &Last & Avg &Last\\
				\midrule
				\multirow{7}*{\tabincell{c}{10k}}
				&iCaRL-cnn \cite{Rebuffi2017iCaRLIC}&31.14 &21.32  &25.61  &17.43 \\
				&iCaRL-ncm \cite{Rebuffi2017iCaRLIC}&42.53  &32.25 &34.83  &27.08 \\
				&UCIR \cite{Hou2019LearningAU}&60.47  &52.25 &56.95  &48.12 \\
				&UCIR-ITO \cite{zhu2023imitating}&62.61  &55.71 &59.92  &51.87 \\
				&PODnet \cite{Douillard2020SmallTaskIL}&63.87  &54.41 &59.19  &48.82 \\
				&PODnet-ITO \cite{zhu2023imitating}&63.65  &55.45 &61.43  &{52.55} \\
				&DER \cite{yan2021dynamically}&64.11  &{57.54} &{61.95}  &{54.09} \\
				\midrule
				\multirow{5}*{\tabincell{c}{20k}}
				&iCaRL-ncm \cite{Rebuffi2017iCaRLIC}&46.89 &39.34 &43.14 &37.25\\
				&UCIR \cite{Hou2019LearningAU}&61.57&52.14 & 56.56 &46.39 \\
				&PODnet \cite{Douillard2020SmallTaskIL}&64.13 &55.21 &59.17  &50.00 \\
				&GeoDL \cite{simon2021learning} &64.46 &56.70  &{62.20}&-- \\
				&Mnemonics \cite{zhu2023imitating}&63.01&54.97 &61.00  &51.26 \\
				&AANets \cite{zhu2023imitating}&{64.85} &{57.14} &61.78 &52.36 \\
				\midrule
				\multirow{1}*{\tabincell{c}{0}} &
				\cellcolor{mygray}PASS++ (Ours) &\cellcolor{mygray}{64.13}  &\cellcolor{mygray}56.37  & \cellcolor{mygray}60.96&\cellcolor{mygray}49.64   \\
				\bottomrule[1.3pt]
		\end{tabular}}
	\end{center}
	\vskip -0.05in
\end{table}

\vspace*{5pt}
\noindent\textbf{Comparison with generative exemplar-based methods.} To avoid saving real data, generative exemplar-based methods \cite{Shin2017ContinualLW, KemkerK18, gao2023ddgr} adopt generative model or model inversion \cite{yin2020dreaming, smith2021always, gao2022r} to generate old pseudo-examples. In Table~\ref{table-4}, we find that our method performs significantly better than recently developed generative exemplar-based methods like ABD \cite{smith2021always} and R-DFCIL \cite{gao2022r}.

\vspace*{5pt}
\noindent\textbf{Experiments on ImageNet-Full.} To verify the effectiveness of our method
on large-scale dataset, we conduct experiments on the larger and
more difficult ImageNet-Full \cite{deng2009imagenet} dataset which includes 1000 classes. The employed backbone is ResNet18 \cite{he2016deep}.
From the results in Table~\ref{table-5}, we find that our method achieves superior performance than popular exemplar-based methods such as iCaRL, UCIR and PODnet in both 10 and 25 incremental phases. In particular, our method is comparable with DER (w/o pruning) in many cases, indicating the proposed dual bias reduction framework is ready for realistic and challenging CIL settings.

\begin{table}[t]
	\caption{CIL performance with pre-trained DeiT-S/16-611 model \cite{touvron2021training, kim2023learnability} model. Baselines results are come from \cite{lin2023class}.}
	\vskip -0.13in
	\label{table-6}
	\begin{center}
		\renewcommand\tabcolsep{5.6pt}
		\newcommand{\tabincell}[2]{\begin{tabular}{@{}#1@{}}#2\end{tabular}}
		\scalebox{1}{
			\renewcommand{\arraystretch}{1.2}
			\begin{tabular}{lcccccc}
				\toprule[1.3pt]	
				\multirow{2}*{\tabincell{c}{\textbf{Dataset}}} &	
				\multicolumn{2}{c}{\makecell{CIFAR100 \\ ($T = 10$)}} & \multicolumn{2}{c}{\makecell{TinyImageNet \\ ($T = 5$)}} & \multicolumn{2}{c}{\makecell{TinyImageNet \\ ($T = 10$)}} \\
				\cmidrule(lr){2-3} \cmidrule(lr){4-5} \cmidrule(lr){6-7}
				& Last &Avg &Last &Avg &Last &Avg \\
				\midrule
				Upper  &82.76 &87.20 &72.52 &77.03 &72.52 &77.03\\
				\midrule
				OWM \cite{zeng2019continual} &21.39 &40.10 &24.55 &45.18 &17.52 &35.75\\
				ADAM \cite{zhou2023revisiting}  &61.21 &72.55 &50.11 &61.85 &49.68 &61.44\\
				PASS \cite{Zhu_2021_CVPR} &68.90 &77.01 &61.03 &67.12 &58.34 &67.33\\
				HATCIL \cite{kimtheoretical} &62.91 &73.99 &59.22 &69.38 &54.03 &65.63\\
				SLDA \cite{hayes2020lifelong}  &67.82 &77.72 &57.93 &66.03 &57.93 &67.39\\
				L2P \cite{wang2022learning} &61.72 &72.88 &59.12 &67.81 &54.09 &64.59 \\
				\rowcolor{mygray}
				PASS++ (Ours) &\textbf{72.12} &\textbf{82.14} &\textbf{66.72} &\textbf{75.61} &\textbf{64.44} &\textbf{74.84} \\
				\midrule
				iCaRL \cite{Rebuffi2017iCaRLIC}  &68.90 &76.50 &53.13 &61.36 &51.88 &63.56\\
				A-GEM \cite{chaudhry2018efficient}  &25.21 &43.83 &30.53 &49.26 &21.90 &39.58\\
				EEIL \cite{castro2018end} &68.08 &81.10 &53.34 &66.63 &50.38 &66.54\\
				GD \cite{lee2019overcoming} &64.36 &80.51 &53.01 &67.51 &42.48 &63.91\\
				DER++ \cite{buzzega2020dark} &69.73 &80.64 &55.84 &66.55 &54.20 &67.14\\
				HAL \cite{chaudhry2021using} &67.17 &77.42 &52.80 &65.31 &55.25 &64.48\\
				DER \cite{yan2021dynamically}  &73.30 &82.89 &59.57 &70.32 &57.18 &70.21\\
				FOSTER \cite{wang2022foster} &71.69 &81.16 &54.44 &69.95 &55.70 &70.00\\
				BEEF \cite{wang2022beef} &72.09 &81.91 &61.41 &71.21 &58.16 &71.16\\
				MORE \cite{kim2022multi} &70.23 &81.24 &64.97 &74.03 &63.06 &72.74\\
				ROW \cite{kim2023learnability}  &74.72 &82.87 &65.11 &74.16 &63.21 &72.91\\
				TPL \cite{lin2023class}  &\textbf{76.53} &\textbf{84.10} &\textbf{68.64} &\textbf{76.77} &\textbf{67.20} &\textbf{75.72}\\
				\bottomrule[1.3pt]
		\end{tabular}}
	\end{center}
	\vskip -0.15in
\end{table}
\vspace*{5pt}\noindent
\textbf{Experiments with pre-trained model.} Recently, some works built the incremental learner based on pre-trained model. We also verify the effectiveness of PASS++ by combining it with pre-trained model via low-rank adaptation \cite{hulora}, as described in Sec. \ref{sec:pretrained}. Specifically, to
prevent information leak \cite{kim2023learnability} in incremental learning, following \cite{kim2023learnability, lin2023class}, we adopt DeiT-S/16 model \cite{touvron2021training} pre-trained using 611 classes of ImageNet after removing 389
classes that overlap with classes in CIFAR and TinyImageNet. Then, the pre-trained network is fixed and LoRA \cite{hulora} modules (we set rank $r=4$) are inserted at each transformer layer and optimized with the overall learning objective in Sec. \ref{sec:objective}. The results in Table \ref{table-6} show that our
PASS++ outperforms prompt-based L2P \cite{wang2022learning} and adapter based ADAM \cite{zhou2023revisiting}, performing the best in both average incremental accuracy and last accuracy among non-exemplar methods. Moreover,
PASS++ outperforms recent strong exemplar-based methods which leverage a replay buffer of 2000 exemplars such as FOSTER \cite{wang2022foster}, BEEF \cite{wang2022foster} , MORE \cite{wang2022foster}  and ROW \cite{wang2022foster}. And only the latest exemplar-based method TPL \cite{lin2023class} performs better than ours. The above results further demonstrate the superiority of our method when integrating with pre-trained models.

\begin{table}[t]
	\caption{The effectiveness of each component in our method. The \textbf{last} accuracy (\%) and forgetting (\%) are reported.}
	\vskip -0.13in
	\label{table-7}
	\begin{center}
		\renewcommand\tabcolsep{1.3pt} 
		\newcommand{\tabincell}[2]{\begin{tabular}{@{}#1@{}}#2\end{tabular}}
		\scalebox{1}{
			\renewcommand{\arraystretch}{1.2}
			\begin{tabular}{llcccccc}
				\toprule[1.3pt]
				\multirow{2}*{\tabincell{c}{\textbf{Metric}}} &
				\multirow{2}*{\tabincell{c}{\textbf{Method}}} &
				\multicolumn{3}{c}{\makecell{\ \textbf{CIFAR-100}}} & \multicolumn{3}{c}{\makecell{\ \textbf{TinyImageNet}}} \\
				\cmidrule(lr){3-5} \cmidrule(lr){6-8}	
				&& $T = 5$ & $T = 10$ & $T = 20$  &$T = 5$  & $T = 10$ & $T = 20$ \\
				\midrule
				\multirow{5}*{\tabincell{c}{Last ACC}} 
				& Baseline & 17.15 & 8.46 & 8.57 & 9.71 & 6.53 & 6.60 \\
				& + protoAug &50.19 & 39.80 & 38.61 & 33.11& 26.52 & 20.97 \\
				& + SST &55.67 & 49.03 & 48.48 & 41.58 & 39.28 & 32.78  \\
				& + Hardness &56.61 &55.26 &51.30 &42.40 &43.20 &37.68 \\
				& + Ensemble &\textbf{59.87}  &\textbf{57.69}  &\textbf{53.43}  &\textbf{46.93}  &\textbf{46.66}  &\textbf{40.53}  \\
				\midrule
				\multirow{4}*{\tabincell{c}{Forgetting }} 
				& + protoAug & 28.72 & 35.70 &40.59 & 20.96&35.33&43.91 \\
				& + SST &25.20 & 30.25&30.61&18.04&23.12&30.55  \\
				& + Hardness &\textbf{22.98} &\textbf{21.94} &\textbf{23.54} &\textbf{19.76} &\textbf{18.62} &\textbf{11.89} \\
				& + Ensemble &23.98 &23.00 &25.42 & 21.24 &21.02 &16.61 \\
				\bottomrule[1.3pt]
		\end{tabular}}
	\end{center}
	\vskip -0.05in
\end{table}

\begin{figure}[t]
	\begin{center}
		\centerline{\includegraphics[width=\columnwidth]{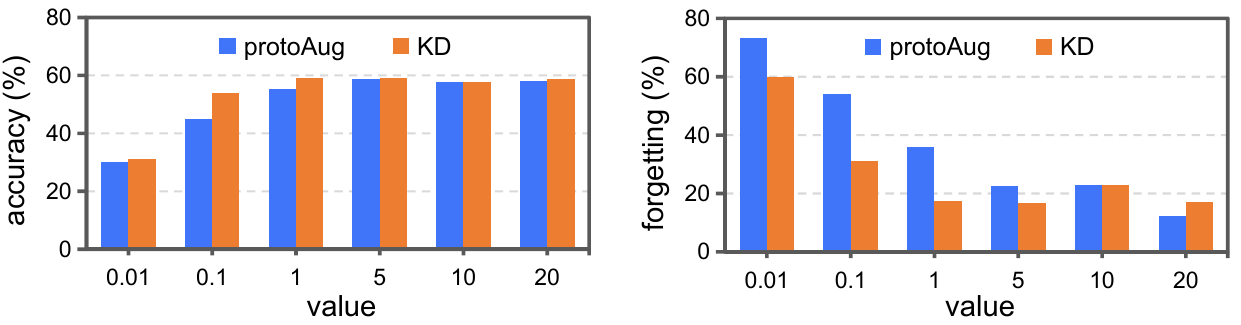}}
		\vskip -0.07in
		\caption{Last accuracy and forgetting with different values of hyper-parameters $\alpha$ (protoAug) and $\beta$ (KD) in Eq.~(\ref{eq11}).}
		\label{figure-8}
	\end{center}
	\vskip -0.25in
\end{figure}

\subsection{Further Analysis and Discussion} \label{sec:analysis}
\subsubsection{Ablation Study and More Results}
\begin{figure*}[t]
	\begin{center}
		\centerline{\includegraphics[width=0.97\textwidth]{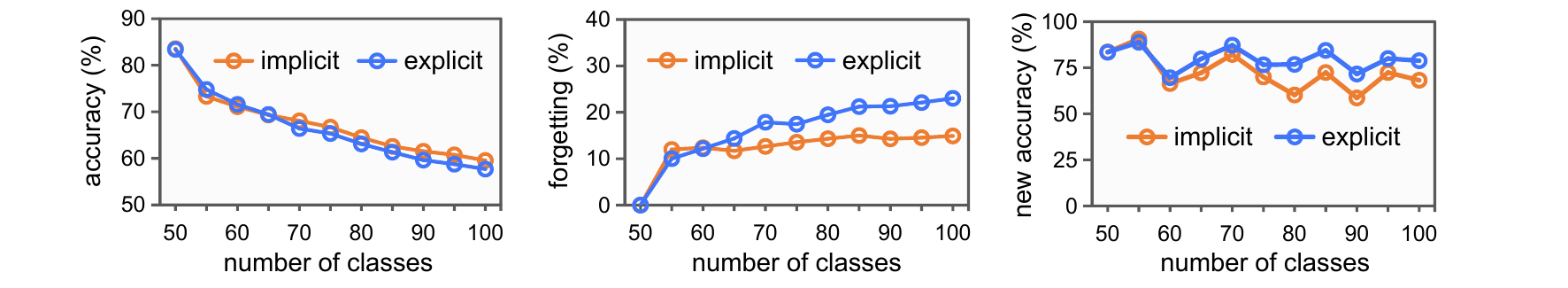}}
		\vskip -0.1in
		\caption{Comparison of explicit and implicit protoAug on CIFAR-100 ($T=10$).}
		\label{figure-9}
	\end{center}
	\vskip -0.2in
\end{figure*}
\noindent\textbf{Each component in PASS++.} We conduct ablation studies to analyze the effect of the individual aspects of PASS++. From the results in
Table~\ref{table-7}, we can observe that:  \textbf{(1)} Baseline which only uses knowledge distillation (KD) completely fails in CIL without protoAug. However, in Fig.~\ref{figure-8}, we have experimentally observed that the performance will drop significantly without KD. As demonstrated in Sec. \ref{sec:objective}, KD is critical for the success of PASS++.
\textbf{(2)} ProtoAug successfully mitigates the imbalance problem and achieves much better results than Baseline, \emph{e.g.}, protoAug improves the performance of Baseline with a margin of $31.34\%$ on CIFAR-100 (10 phases).
\textbf{(3)} The performance of protoAug could be significantly improved by SST, which indicates that SSL and protoAug could benefit from each other. 
\textbf{(4)} The hardness-aware protoAug and prediction ensemble further improve the performance by learning feature instances near the decision boundary and leveraging multi-view information, respectively. 

\vspace*{5pt}
\noindent\textbf{Hyperparameter.} Fig.~\ref{figure-8} shows how the strength of protoAug and KD affect the performance of the incremental learner in terms of final accuracy and forgetting. As can be seen, larger strength of protoAug and KD leads to improved accuracy and less forgetting (better stability), which is predictable due to the stronger regularization effect. However, further increasing the strength would lead to lower new task accuracy (worse plasticity). Therefore, to achieve overall accuracy, the weights of protoAug and KD are set to be 10 in our experiments.

\begin{figure}[h]
	\begin{center}
		\centerline{\includegraphics[width=\columnwidth]{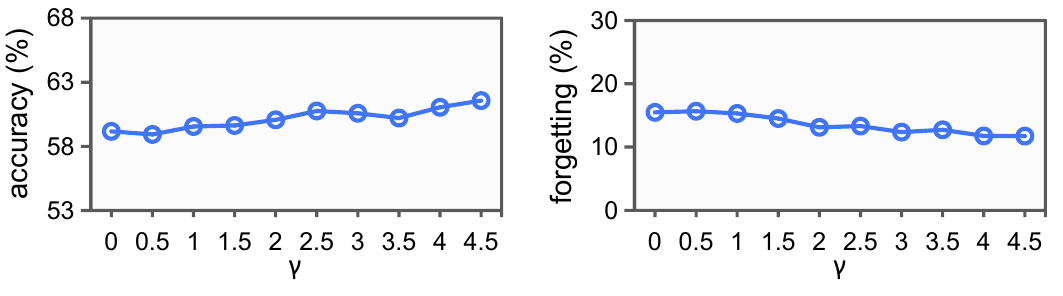}}
		\vskip -0.12in
		\caption{Influence of the hyper-parameter ($\gamma$ in Eq.~(\ref{eq10})) in implicit protoAug. CIFAR-100 ($T=10$).}
		\label{figure-100}
	\end{center}
	\vskip -0.23in
\end{figure}

\vspace*{5pt}
\noindent\textbf{Explicit protoAug v.s. implicit protoAug.}
As demonstrated in Sec. \ref{sec:protoAug}, both explicit protoAug and implicit protoAug aim to reduce the classifier bias by maintaining the decision boundary of previously learned classes. Different from explicit protoAug which generates pseudo-instances of old classes explicitly, implicit protoAug optimizes an upper bound of the loss which is further transformed into a regularization term. Fig.~\ref{figure-9} compares explicit and implicit protoAug on CIFAR-100 (the hardness-aware protoAug is also applied for implicit protoAug here). Those two strategies have similar incremental accuracy. However, since implicit protoAug generates infinite feature instances for old classes, the regularization effect is stronger. Consequently, the forgetting of old classes is reduced, but the new task accuracy is also slightly worse. In implicit protoAug, $\gamma$ in Eq.~(\ref{eq10}) is a hyper-parameter that influences the variance of generated features of old classes. Particularly, only the prototypes are used for knowledge retention when $\gamma=0$. Fig.~\ref{figure-100} reports the results of how the parameter influences the performance of our method, and the value of $\gamma$ should not be too small or too large. 

\vspace*{5pt}
\noindent\textbf{Experiments with equally divided classes in all tasks.} In Table \ref{table-2}, Table \ref{table-3}, Table \ref{table-4} and Table \ref{table-5}, the first task in all experiments contains about half of the classes in the dataset. Table \ref{table-6} reports the results of \emph{tasks with equal classes} with pre-trained model. Table \ref{table-8} reports the setting where each task contains equal classes without pre-trained model. Specifically, we divide each dataset into five tasks equally and the results show that our method performs better or is comparable with data replay method PODNet \cite{Douillard2020SmallTaskIL}. Since DER \cite{yan2021dynamically} is a strong data replay method with continually expandable feature extractor backbones, it could outperform our exemplar-free approach in this CIL setting.

\begin{table}[t]
	\vskip 0.05in
	\caption{CIL with equally divided classes in all tasks.}
	\vskip -0.15 in
	\label{table-8}
	\begin{center}
		\renewcommand\tabcolsep{5.5pt}
		\newcommand{\tabincell}[2]{\begin{tabular}{@{}#1@{}}#2\end{tabular}}
		\scalebox{1}{
			\renewcommand{\arraystretch}{1.2}
			\begin{tabular}{lcccccccccc}
				\toprule[1.3pt]	
				\multirow{2}*{\tabincell{c}{\textbf{Dataset}}} &	
				\multicolumn{2}{c}{\makecell{CIFAR-100}} & \multicolumn{2}{c}{\makecell{TinyImageNet}} & \multicolumn{2}{c}{\makecell{ImageNet-Subset}}\\
				\cmidrule(lr){2-3} \cmidrule(lr){4-5} \cmidrule(lr){6-7}
				& Last &Avg &Last &Avg &Last &Avg \\
				\midrule
				PODnet \cite{Douillard2020SmallTaskIL} &45.58 &58.31 &36.36 &47.04 &49.76 &64.72 \\
				DER \cite{yan2021dynamically} &54.83 &63.58 &42.58 &52.36 &65.88 &74.54 \\
				\rowcolor{mygray}
				\midrule
				PASS++ &52.21 &65.15 &37.19 &49.32 &48.75 &63.95\\
				\bottomrule[1.3pt]
		\end{tabular}}
	\end{center}
	\vskip -0.05in
\end{table}

\begin{figure}[t]
	\begin{center}
		\centerline{\includegraphics[width=\columnwidth]{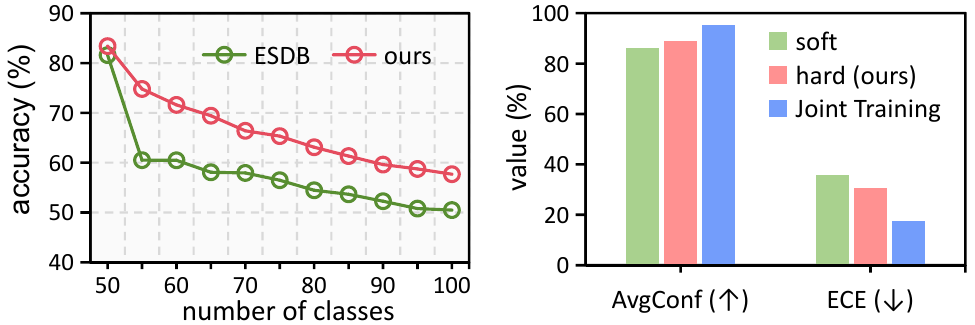}}
		\vskip -0.07in
		\caption{Hard v.s. soft label in hardness-aware protoAug.}
		\label{figure-10}
	\end{center}
	\vskip -0.25in
\end{figure}

\begin{table*}[t]
	\vskip -0.05in
	\caption{Experimental comparison of incremental accuracy between vanilla ensemble and our class node expansion ensemble. CIFAR-100 with 10 incremental stages.}
	\vskip -0.15in
	\label{table-a2}
	\begin{center}
		\renewcommand\tabcolsep{6pt}
		\begin{small}
			\newcommand{\tabincell}[2]{\begin{tabular}{@{}#1@{}}#2\end{tabular}}
			\scalebox{0.93}{
				\renewcommand{\arraystretch}{1.1}
				\begin{tabular}{l|ccccccccccc}
					\toprule[1.5pt]
					Method \textbackslash ~Stage & 0 & 1 & 2 & 3 & 4 & 5 & 6 &7 &8 &9 &10 \\
					\midrule
					PODNet & \textbf{82.14} & \textbf{76.33} & \textbf{71.32} & \textbf{66.40} & \textbf{61.53} & \textbf{58.03} & \textbf{55.20} & \textbf{53.19} & \textbf{51.17} & \textbf{50.61} & \textbf{49.45} \\
					\emph{w}/ vanilla ensemble & 80.82 & 71.11 & 56.57 & 49.60 & 44.61 & 40.52 & 37.38 & 36.52 & 36.72 & 38.61 & 37.09 \\
					\midrule
					DER & \textbf{80.86} & \textbf{76.09} & \textbf{73.93} & \textbf{71.22} & \textbf{68.64} & \textbf{65.21} & \textbf{62.58} & \textbf{60.72} & \textbf{59.14} & \textbf{58.32} & \textbf{57.38} \\
					\emph{w}/ vanilla ensemble & 75.02 & 59.58 & 57.47 & 49.82 & 49.37 & 44.48 & 42.72 & 39.55 & 39.47 & 38.68 & 37.86 \\
					\midrule
					PASS + Hardness & 79.46 & 71.29 & 68.47 & 66.03 & 64.96 & 63.28 & 60.98 & 59.46 & 58.31 & 56.76 & 55.26 \\
					\emph{w}/ vanilla ensemble & 77.26 & 51.18 & 53.43 & 49.83 & 49.96 & 50.41 & 47.06 & 47.11 & 46.11 & 45.85 & 44.69 \\
					\emph{w}/ our ensemble & \textbf{83.40} & \textbf{74.80} & \textbf{71.60} & \textbf{69.43} & \textbf{66.40} & \textbf{65.33} & \textbf{63.10} & \textbf{61.33} & \textbf{59.62} & \textbf{58.75} & \textbf{57.69}\\
					\bottomrule[1.5pt]
			\end{tabular}}
		\end{small}
	\end{center}
	\vskip -0.1in
\end{table*}
\vspace*{5pt}
\noindent
\textbf{Hard v.s. soft label in hardness-aware protoAug.} As described in Sec. \ref{sec:haprotoAug}, we select the nearest sample of new classes to mix with old class prototypes in the final deep feature space and assign the \emph{hard} label of the corresponding old class for the mixed data. A recent exemplar-based approach ESDB mixs the raw samples of new and old classes and assigns \emph{soft} label. Despite that soft label is effective for exemplar-based methods in \cite{li2024esdb}, it has negative effectiveness in our non-exemplar framework. 
To demonstrate this point, Fig. \ref{figure-10} compares the effectiveness of hard label in our method and soft label in ESDB \cite{li2024esdb}. As can be observed, the CIL performance would drop a lot when assigning soft label for mixed data.  
To understand the above results, we provide an explanation from the perspective of confidence \textbf{calibration} \cite{guo2017calibration} by measuring the Expected
Calibration Error (ECE) \cite{Naeini2015ObtainingWC}. Specifically, we compare the CIL final models trained on CIFAR-100 after learning 10 incremental stages using the soft and hard label mixing strategies with the Joint Training model, respectively. The results are shown in Fig. \ref{figure-10}, where we can observe that the calibration of CIL model trained with hard label mixing data is more aligned with Joint Training, which explains the reason why our hardness-aware protoAug is more effective than the soft label mixing strategy used in \cite{li2024esdb} for CIL.

\vspace*{5pt}
\noindent
\textbf{Multi-view ensemble v.s. vanilla rotation ensemble.} 
In our method, the number of class nodes in the classifier is expended, extending the original $k$ classes to 4$k$ classes, i.e., each view $\delta$ shares $k$ class nodes $g_{\bm{\varphi}, \delta}$. The proposed multi-view ensemble aggregates the logits computed on the corresponding $k$ classes of each view at inference time. This is quite different from the vanilla ensemble, where each view of an image is fed into the classifier $g_{\bm{\varphi}}$ with the same $k$ class nodes. 
We conduct experiments to compare the two ensemble strategies in Table \ref{table-a2}. We can draw two key observations: (1) In our method, if we fed the 4 rotating views to the original classifier with $k=100$ class nodes and then performed vanilla ensemble by simply averaging outputs of 4 rotating views, the CIL performance drop a lot, e.g., the final accuracy drop from 55.26\% to 44.69\% in Table \ref{table-a2}. On the contrary, the proposed multi-view ensemble has remarkable positive effectiveness for CIL. This is because the SST has guided the model to learn discriminative representations among different rotation angles during training, and the extracted corresponding class nodes for each angle $\theta \in \{90, 180, 270\}$ successfully capture the corresponding discriminative information.  (2) When applying the vanilla ensemble on strong exemplar-based CIL methods such as PODNet \cite{Douillard2020SmallTaskIL} and DER \cite{yan2021dynamically}, we observe that simply averaging output of 4 rotating views has negative effectiveness. This is reasonable because such a strong rotation degree, e.g., $\theta \in \{90, 180, 270\}$ leads to noisy inputs at inference time, disturbing the original logits when averaging them.

\vspace*{5pt}
\noindent
\textbf{Training and inference time with SST.} Our method involves SST in training and multi-view ensemble in inference.
Table \ref{table-9} provides the statistics of training and inference time with and without SST. It is shown that the time of training and inference with SST is less than twice that of without SST.

\subsubsection{On the Plasticity of Incremental Learner}
In our method, to keep the plasticity for learning new knowledge, the feature extractor is updated on new tasks continually, which adapts and consolidates representation knowledge of new classes. To demonstrate the necessity of updating the feature extractor, we compare our method with a strong baseline that freezes the feature extractor and applies NCM classifier in incremental phases. In this way, the prototypes of old classes keep unchanged and it is identical to the NCM classifier with all data as memory.

\begin{table}[t]
	\vskip 0.05in
	\caption{Training and inference time (in seconds).}
	\vskip -0.15in
	\label{table-9}
	\begin{center}
		\renewcommand\tabcolsep{6.5pt}
		\newcommand{\tabincell}[2]{\begin{tabular}{@{}#1@{}}#2\end{tabular}}
		\scalebox{1}{
			\renewcommand{\arraystretch}{1.2}
			\begin{tabular}{lrrrr}
				\toprule[1.3pt]	
				\multirow{2}*{\tabincell{c}{\textbf{Dataset}}} &	
				\multicolumn{2}{c}{\makecell{Training (1 epoch)}} & \multicolumn{2}{c}{\makecell{Inference}}\\
				\cmidrule(lr){2-3} \cmidrule(lr){4-5}
				& w/o SST  &w/ SST  &w/o SST &w/ SST \\
				\midrule
				CIFAR-100  &21.95 &29.28  &3.85 &6.17 \\
				TinyImageNet  &46.80 &72.14     &8.06 &16.02 \\
				ImageNet-Subset  &102.35 &156.93    &13.15 &22.94 \\
				\bottomrule[1.3pt]
		\end{tabular}}
	\end{center}
	\vskip -0.1in
\end{table}

\begin{figure}[t]
	\begin{center}
		\centerline{\includegraphics[width=\columnwidth]{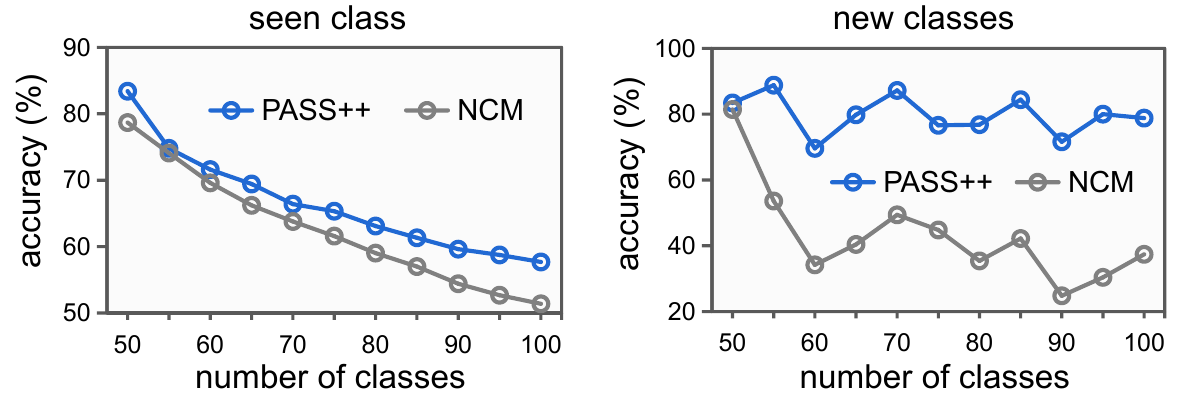}}
		\vskip -0.1in
		\caption{Comparison to NCM classifier with fixed feature extractor on CIFAR-100 ($T=10$).}
		\label{figure-12}
	\end{center}
	\vskip -0.2in
\end{figure}
Comparative results Fig.~\ref{figure-12} show that NCM with fixed feature extractor achieves good accuracy on seen classes but quite low accuracy on new classes in each incremental phase. On the one hand, the model lacks plasticity for new classes, making it hard to distinguish them from each other. On the other hand, the new and old classes are overlapped in deep feature space, making many inputs from new classes misclassified as old classes. Therefore, the new class accuracy is quite low. We also find that although the recent method SSRE \cite{zhu2022self} yields notable average incremental accuracy, its new class accuracy is also low: on CIFAR-100 ($T=10$), the accuracy of final task (includes 5 classes) is only 44.32\%, while ours is 78.80\%. In summary, compared with NCM and SSRE \cite{zhu2022self}, our method not only performs better on seen classes but also achieves much higher accuracy on new classes, indicating better stability and plasticity. 
Interestingly, a recent work \cite{kim2023stability} reveals that many CIL methods heavily favor stability over plasticity, yielding low accuracy for new classes in each incremental phase. Here, we would like to emphasize that an incremental learner should be plastic enough to learn concepts from new classes and be stable to retain knowledge learned from previously seen classes.

\subsubsection{Discussion of Covariance and Memory Cost}
\begin{figure}[t]
	\begin{center}
		\centerline{\includegraphics[width=0.9\columnwidth]{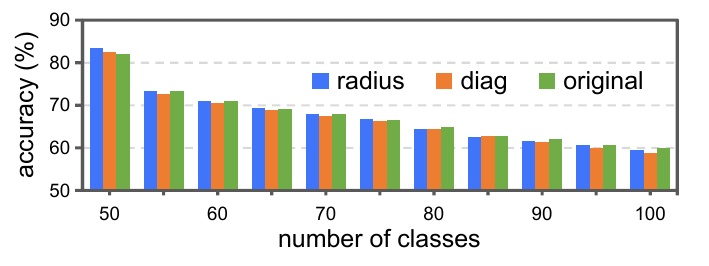}}
		\vskip -0.1in
		\caption{Different form of covariance matrix used for implicit prototype augmentation. CIFAR-100 ($T=10$).}
		\label{figure-13}
	\end{center}
	\vskip -0.25in
\end{figure}

In our main experiments, to reduce memory usage in the incremental process, we only save a class mean vector in the deep feature space for each class and a shared radius (a scalar value) for all old classes. This choice is based on the assumption that each class obeys a spherical Gaussian distribution. Alternative ways could be storing the original or diagonal covariance matrix for each old class.
As shown in Fig.~\ref{figure-13}, using the original covariance matrix is slightly better than the diagonal and spherical form. However, storing the original covariance matrix might be inefficient when the matrix dimension is large. Therefore, we use the radius form to significantly reduce the memory requirement.

\vspace*{5pt}
\noindent\textbf{Memory cost.} To quantitatively compare the memory cost of DBR (prototype along with diagonal covariance or radius scale) and exemplar-based method, we use an ``entry'' to denote a value required for each dataset. Table~\ref{table-10} presents comparison between exemplar-based methods and our DBR. It can be observed that exemplar-based methods (\emph{e.g.}, iCaRL, UCIR, DER) require the largest memory. For ImageNet-Subset, saving 1000 images of 224$\times$224$\times$3 requires 150.5M entries; our method only requires 50K entries, which are negligible compared with exemplar-based method. 

\begin{table}[t]
	\vskip 0.05in
	\caption{Comparison of memory cost (in entries).}
	\vskip -0.15in
	\label{table-10}
	\begin{center}
		\renewcommand\tabcolsep{8.7pt}
		\newcommand{\tabincell}[2]{\begin{tabular}{@{}#1@{}}#2\end{tabular}}
		\scalebox{1}{
			\renewcommand{\arraystretch}{1.25}
			\begin{tabular}{lcccc}
				\toprule[1.3pt]	
				\multirow{2}*{\tabincell{c}{\textbf{Dataset}}} &	
				\multicolumn{2}{c}{\makecell{Exemplar-based}} & \multicolumn{2}{c}{\makecell{PASS++ (ours)}}\\
				\cmidrule(lr){2-3} \cmidrule(lr){4-5}
				& $R=20$  &$R=10$  &diag &\cellcolor{mygray}radius \\
				\midrule
				CIFAR-100 &6.1M &3.0M  &0.1M    &\cellcolor{mygray}50K \\
				TinyImageNet  &49.2M &24.6M     &0.2M &\cellcolor{mygray}0.1M \\
				ImageNet-Subset  &301.1M &150.5M    &0.1M &\cellcolor{mygray}50K \\
				ImageNet-Full  &3010.6M &1505.3M     &1.0M &\cellcolor{mygray}0.5M \\
				\bottomrule[1.3pt]
		\end{tabular}}
	\end{center}
	\vskip -0.15in
\end{table}

\subsubsection{Evaluation Incremental Model under Distribution Shift}
\begin{figure}[h]
	\begin{center}
		\vskip -0.05in
		\centerline{\includegraphics[width=1.03\columnwidth]{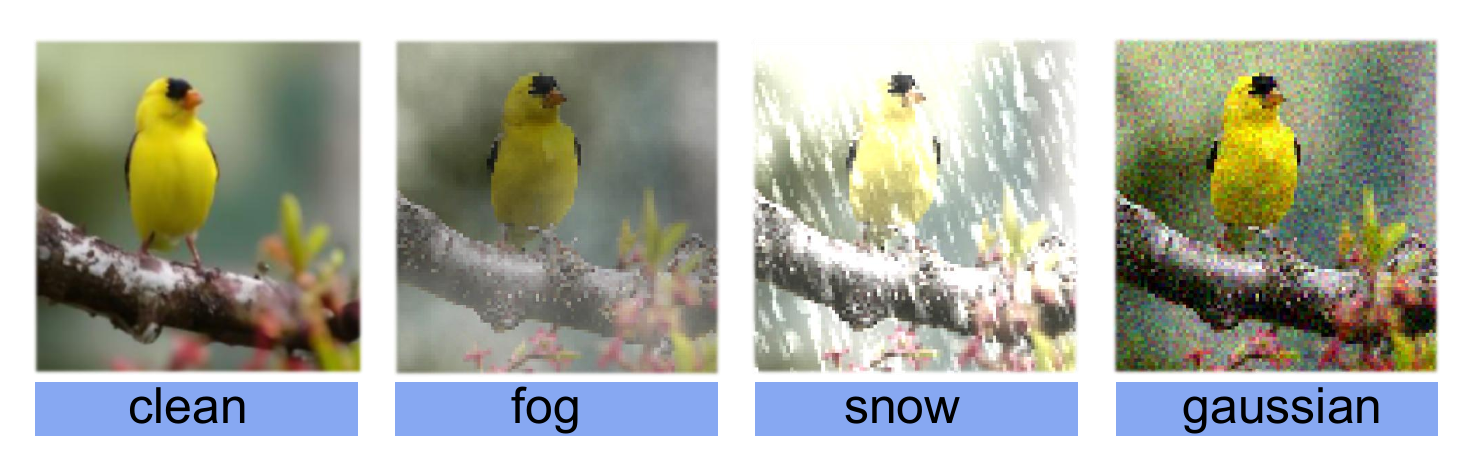}}
		\vskip -0.1in
		\caption{Visualization of several types of distribution shifts.}
		\label{figure-14}
	\end{center}
	\vskip -0.25in
\end{figure}

In practical applications, environments can be easily changed, \emph{e.g.}, weather change from sunny to cloudy then to rainy \cite{zhu2022rethinking}. The model still needs to make accurate prediction under such distribution or domain shift conditions. However, existing CIL works assume that the training and testing samples of each task come from the same distribution, overlooking the model's robustness to unknown distribution shifts. In this paper, we argue that it is necessary to evaluate the model's performance under distribution drifts. Fig.~\ref{figure-14} shows a clean image can encounter various distribution shifts in practice. Here we conduct empirical evaluation to reflect the robustness of incremental learning models under distribution drift. Specifically, the model is trained on CIFAR-100 (10 phases) incrementally and evaluated on the CIFAR-100-C dataset \cite{hendrycksbenchmarking}, which contains copies of the original
validation set with 15 types of corruptions of algorithmically generated corruptions from noise, blur, weather, and digital
categories. Each type of corruption has five different severity levels, and we select the mildest level of distribution drift for testing.

\begin{figure}[h]
	\vskip -0.1in
	\begin{center}
		\centerline{\includegraphics[width=\columnwidth]{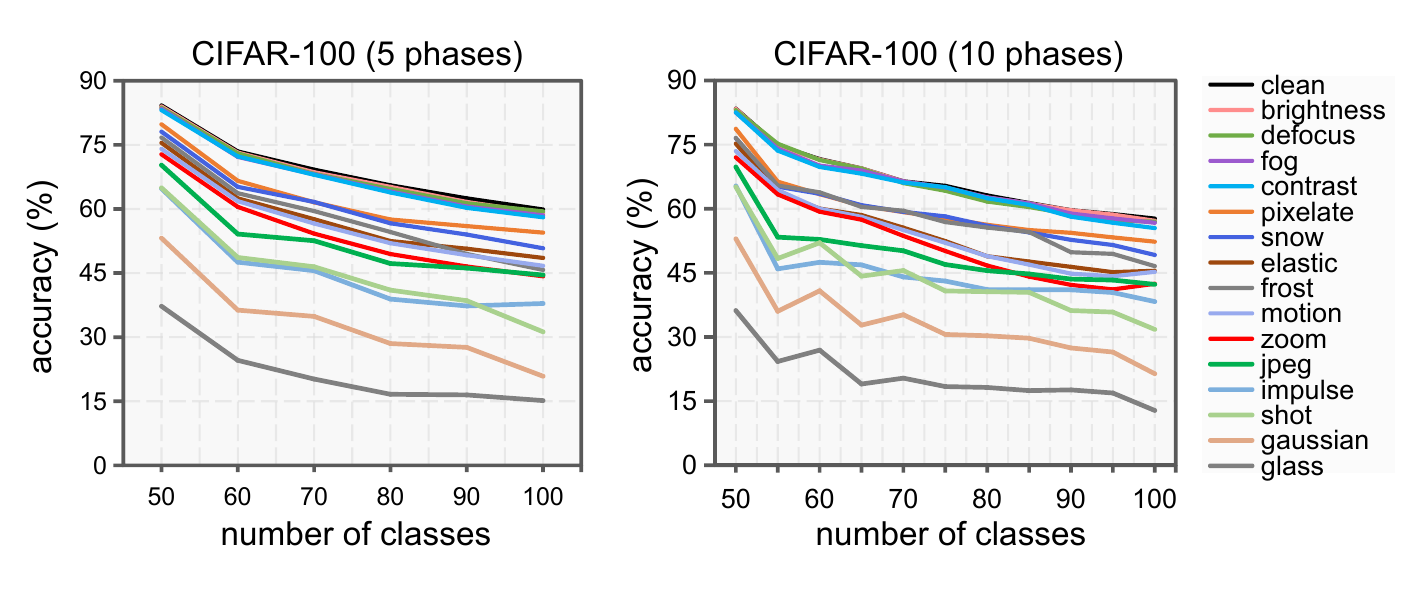}}
		\vskip -0.15in
		\caption{Incremental accuracy curves of our method under different types of distribution shifts.}
		\label{figure-15}
	\end{center}
	\vskip -0.23in
\end{figure}

\begin{figure}[h]
	\vskip -0.1in
	\begin{center}
		\centerline{\includegraphics[width=1.07\columnwidth]{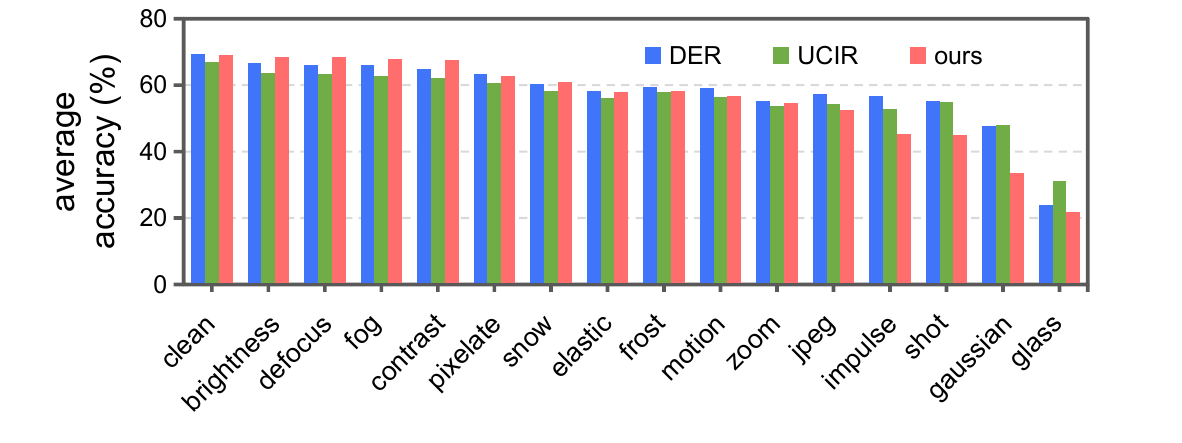}}
		\vskip -0.15in
		\caption{Comparison results under distribution shifts.}
		\label{figure-16}
	\end{center}
	\vskip -0.25in
\end{figure}

The step-wise incremental accuracy curves of our method are shown in Fig.~\ref{figure-15}, and the average incremental accuracy results of our approach and exemplar-based methods like UCIR \cite{Hou2019LearningAU} and DER \cite{yan2021dynamically} under 15 different corruption scenarios are summarized in Fig.~\ref{figure-16}. From those results, we have two key observations: \textbf{(1)} \textbf{\emph{First,}} compared to the results on the clean testset, the models suffer from performance degradation under all types of corruptions, indicating that even the state-of-the-art CIL models are vulnerable to distribution drift during testing. \textbf{(2)} \textbf{\emph{Second,}} the models demonstrate better robustness to distribution drifts such as brightness, snow, and fog, while their performance significantly decreases in distribution drift like Gaussian noise. For instance, DER \cite{yan2021dynamically} achieves an average accuracy of 66.73\% on the clean test set, whereas it only achieves 47.79\% on the test set with Gaussian noise corruption. These results indicate that existing incremental learning models are highly unreliable in high-risk and safety-critical applications due to distribution drift. Understanding the performance of incremental learners under distribution drift and improving their robustness is crucial when deploying them in real-world scenarios.

\subsubsection{ProtoAug v.s. Data Replay}
In Sec. \ref{sec:main-experiment}, we have shown that our method can achieve strong performance without strong any raw data of old classes during the course of CIL. Considering the dominant role of exemplar-based methods in CIL, we provide a detailed discussion between data replay and our method.

\vspace*{3pt}
\noindent
\textbf{Privacy issue.} Privacy concerns is important in some real-world applications involving personalized data, \emph{e.g.}, in the field of healthcare and security \cite{li2020federated}. For CIL, exemplar-based methods often store a fraction of old raw data for future learning, which would be unsuitable for privacy-sensitive applications. Contrarily, the proposed protoAug protects privacy at the basic level because it does not store any raw data during the course of CIL, but estimates the class distribution information in the deep feature space, which is considered to be privacy-preserving \cite{luo2021no}. We would like to highlight that developing privacy-preserving learning approaches is crucial for future CIL research.

\vspace*{3pt}
\noindent
\textbf{Memory limitation.} A basic assumption in CIL is the memory restrictions that the model can not store old data for joint training when learning new classes. Exemplar-based methods suffer from memory limitation for long-step incremental learning and applications with limited computational budget \cite{lesort2020continual}. While our method is quite memory-efficient (Table~\ref{table-10}), and not affected by image size.

\vspace*{3pt}
\noindent
\textbf{Training efficiency.} In data replay based methods, the memory storage is increasing and the model has to relearn the stored exemplars in each following task, which is inefficient for long-step incremental applications. Moreover, the model could easily overfit stored old exemplars if only storing a few samples for each class. 

\vspace*{3pt}
\noindent
\textbf{Human-likeness.} Humans and other animals excel at incremental learning throughout their lifespan \cite{wixted2004psychology}. In some way, directly storing some old training samples is less human-like from the biological perspective \cite{kumaran2016learning}. 
In our method, the prototype of each old class
is used to simulate the abstract memories of the class in human brain \cite{biederman1985human}.

\section{Conclusion} \label{sec:conclusion}
Existing CIL approaches are mostly based on exemplar replay.
In this paper, we propose a simple and effective non-exemplar method named PASS++ based on dual bias reduction, which consists of self-supervised transformation in input space and prototype augmentation in deep feature space. We extend the proposed framework to pre-trained backbone based on low-rank adaptation \cite{hulora}.
Without storing any old training samples, PASS++ is capable of alleviating the catastrophic forgetting problem in CIL, performing comparably with state-of-the-art exemplar-based approaches under different settings. Towards real-world applications, we call to evaluate incremental model under distribution shift. 
Finally, we discuss the advantages of our method compared with exemplar-based methods, with the hope of drawing the attention of researchers back to non-exemplar methods. 
Future works will consider theoretical analysis of our framework and more challenging scenarios like semi-supervised and unsupervised CIL.

{\small
	\bibliographystyle{unsrt}
	\bibliography{refer}

\begin{thebibliography}{100}

\bibitem{wixted2004psychology}
John~T Wixted.
\newblock The psychology and neuroscience of forgetting.
\newblock {\em Annual Review of Psychology}, 55:235--269, 2004.

\bibitem{tulving197212}
Endel Tulving.
\newblock Episodic and semantic memory.
\newblock {\em Organization of Memory}, pages 381--403, 1972.

\bibitem{hadsell2020embracing}
Raia Hadsell, Dushyant Rao, Andrei~A Rusu, and Razvan Pascanu.
\newblock Embracing change: Continual learning in deep neural networks.
\newblock {\em Trends in Cognitive Sciences}, 2020.

\bibitem{Goodfellow2014AnEI}
Ian~J Goodfellow, Mehdi Mirza, Da~Xiao, Aaron Courville, and Yoshua Bengio.
\newblock An empirical investigation of catastrophic forgetting in
  gradient-based neural networks.
\newblock {\em arXiv preprint arXiv:1312.6211}, 2013.

\bibitem{McCloskey1989CatastrophicII}
M.~McCloskey and N.~J. Cohen.
\newblock Catastrophic interference in connectionist networks: The sequential
  learning problem.
\newblock {\em Psychology of Learning and Motivation}, 24:109--165, 1989.

\bibitem{Kirkpatrick2017OvercomingCF}
J.~Kirkpatrick, Razvan Pascanu, Neil~C. Rabinowitz, et~al.
\newblock Overcoming catastrophic forgetting in neural networks.
\newblock {\em Proceedings of the National Academy of Sciences}, 114:3521 --
  3526, 2017.

\bibitem{Li2018LearningWF}
Zhizhong Li and Derek Hoiem.
\newblock Learning without forgetting.
\newblock {\em IEEE Transactions on Pattern Analysis and Machine Intelligence},
  40:2935--2947, 2018.

\bibitem{Zenke2017ContinualLT}
Friedemann Zenke, Ben Poole, and S.~Ganguli.
\newblock Continual learning through synaptic intelligence.
\newblock In {\em International Conference on Machine Learning}, pages
  3987--3995, 2017.

\bibitem{masana2022class}
Marc Masana, Xialei Liu, Bart{\l}omiej Twardowski, Mikel Menta, Andrew~D
  Bagdanov, and Joost Van De~Weijer.
\newblock Class-incremental learning: survey and performance evaluation on
  image classification.
\newblock {\em IEEE Transactions on Pattern Analysis and Machine Intelligence},
  45(5):5513--5533, 2022.

\bibitem{wang2024comprehensive}
Liyuan Wang, Xingxing Zhang, Hang Su, and Jun Zhu.
\newblock A comprehensive survey of continual learning: theory, method and
  application.
\newblock {\em IEEE Transactions on Pattern Analysis and Machine Intelligence},
  2024.

\bibitem{zhou2023deep}
Da-Wei Zhou, Qi-Wei Wang, Zhi-Hong Qi, Han-Jia Ye, De-Chuan Zhan, and Ziwei
  Liu.
\newblock Deep class-incremental learning: A survey.
\newblock {\em arXiv preprint arXiv:2302.03648}, 2023.

\bibitem{Zhu_2021_CVPR}
Fei Zhu, Xu-Yao Zhang, Chuang Wang, Fei Yin, and Cheng-Lin Liu.
\newblock Prototype augmentation and self-supervision for incremental learning.
\newblock In {\em Proceedings of the IEEE/CVF Conference on Computer Vision and
  Pattern Recognition}, pages 5871--5880, 2021.

\bibitem{douillard2022dytox}
Arthur Douillard, Alexandre Ram{\'e}, Guillaume Couairon, and Matthieu Cord.
\newblock Dytox: Transformers for continual learning with dynamic token
  expansion.
\newblock In {\em Proceedings of the IEEE/CVF Conference on Computer Vision and
  Pattern Recognition}, pages 9285--9295, 2022.

\bibitem{Rebuffi2017iCaRLIC}
Sylvestre-Alvise Rebuffi, A.~Kolesnikov, Georg Sperl, and Christoph~H. Lampert.
\newblock icarl: Incremental classifier and representation learning.
\newblock In {\em Proceedings of the IEEE/CVF Conference on Computer Vision and
  Pattern Recognition}, pages 5533--5542, 2017.

\bibitem{Douillard2020SmallTaskIL}
Arthur Douillard, Matthieu Cord, Charles Ollion, et~al.
\newblock Podnet: Pooled outputs distillation for small-tasks incremental
  learning.
\newblock In {\em Proceedings of the European Conference on Computer Vision},
  pages 86--102, 2020.

\bibitem{liu2021adaptive}
Yaoyao Liu, Bernt Schiele, and Qianru Sun.
\newblock Adaptive aggregation networks for class-incremental learning.
\newblock In {\em Proceedings of the IEEE/CVF Conference on Computer Vision and
  Pattern Recognition}, pages 2544--2553, 2021.

\bibitem{yan2021dynamically}
Shipeng Yan, Jiangwei Xie, and Xuming He.
\newblock Der: Dynamically expandable representation for class incremental
  learning.
\newblock In {\em Proceedings of the IEEE/CVF Conference on Computer Vision and
  Pattern Recognition}, pages 3014--3023, 2021.

\bibitem{wang2022foster}
Fu-Yun Wang, Da-Wei Zhou, Han-Jia Ye, and De-Chuan Zhan.
\newblock Foster: Feature boosting and compression for class-incremental
  learning.
\newblock In {\em Proceedings of the European Conference on Computer Vision},
  2022.

\bibitem{nie2023bilateral}
Xing Nie, Shixiong Xu, Xiyan Liu, Gaofeng Meng, Chunlei Huo, and Shiming Xiang.
\newblock Bilateral memory consolidation for continual learning.
\newblock In {\em Proceedings of the IEEE/CVF Conference on Computer Vision and
  Pattern Recognition}, pages 16026--16035, 2023.

\bibitem{wen2023optimizing}
Haitao Wen, Haoyang Cheng, Heqian Qiu, Lanxiao Wang, Lili Pan, and Hongliang
  Li.
\newblock Optimizing mode connectivity for class incremental learning.
\newblock In {\em International Conference on Machine Learning}, 2023.

\bibitem{lin2023class}
Haowei Lin, Yijia Shao, Weinan Qian, Ningxin Pan, Yiduo Guo, and Bing Liu.
\newblock Class incremental learning via likelihood ratio based task
  prediction.
\newblock {\em arXiv preprint arXiv:2309.15048}, 2023.

\bibitem{kim2023learnability}
Gyuhak Kim, Changnan Xiao, Tatsuya Konishi, and Bing Liu.
\newblock Learnability and algorithm for continual learning.
\newblock In {\em International Conference on Machine Learning}, 2023.

\bibitem{parisi2019continual}
German~I Parisi, Ronald Kemker, Jose~L Part, et~al.
\newblock Continual lifelong learning with neural networks: A review.
\newblock {\em Neural Networks}, 113:54--71, 2019.

\bibitem{li2020federated}
Tian Li, Anit~Kumar Sahu, Ameet Talwalkar, and Virginia Smith.
\newblock Federated learning: Challenges, methods, and future directions.
\newblock {\em IEEE Signal Processing Magazine}, 37(3):50--60, 2020.

\bibitem{kumaran2016learning}
Dharshan Kumaran, Demis Hassabis, and James~L McClelland.
\newblock What learning systems do intelligent agents need? complementary
  learning systems theory updated.
\newblock {\em {Trends in Cognitive Sciences}}, 20(7):512--534, 2016.

\bibitem{Shin2017ContinualLW}
Hanul Shin, Jung~Kwon Lee, Jaehong Kim, and Jiwon Kim.
\newblock Continual learning with deep generative replay.
\newblock In {\em Advances in Neural Information Processing Systems}, pages
  2990--2999, 2017.

\bibitem{Wu2018MemoryRG}
Chenshen Wu, L.~Herranz, X.~Liu, Y.~Wang, Joost van~de Weijer, and B.~Raducanu.
\newblock Memory replay gans: Learning to generate new categories without
  forgetting.
\newblock In {\em Advances in Neural Information Processing Systems}, 2018.

\bibitem{gao2022r}
Qiankun Gao, Chen Zhao, Bernard Ghanem, and Jian Zhang.
\newblock R-dfcil: Relation-guided representation learning for data-free class
  incremental learning.
\newblock In {\em Proceedings of the European Conference on Computer Vision},
  pages 423--439. Springer, 2022.

\bibitem{lin2022towards}
Guoliang Lin, Hanlu Chu, and Hanjiang Lai.
\newblock Towards better plasticity-stability trade-off in incremental
  learning: a simple linear connector.
\newblock In {\em Proceedings of the IEEE/CVF Conference on Computer Vision and
  Pattern Recognition}, pages 89--98, 2022.

\bibitem{wang2021training}
Shipeng Wang, Xiaorong Li, Jian Sun, and Zongben Xu.
\newblock Training networks in null space of feature covariance for continual
  learning.
\newblock In {\em Proceedings of the IEEE/CVF Conference on Computer Vision and
  Pattern Recognition}, pages 184--193, 2021.

\bibitem{kong2022balancing}
Yajing Kong, Liu Liu, Zhen Wang, and Dacheng Tao.
\newblock Balancing stability and plasticity through advanced null space in
  continual learning.
\newblock In {\em Proceedings of the European Conference on Computer Vision},
  pages 219--236. Springer, 2022.

\bibitem{lintrgp}
Sen Lin, Li~Yang, Deliang Fan, and Junshan Zhang.
\newblock Trgp: Trust region gradient projection for continual learning.
\newblock In {\em International Conference on Learning Representations}, 2022.

\bibitem{van2022three}
Gido~M van~de Ven, Tinne Tuytelaars, and Andreas~S Tolias.
\newblock Three types of incremental learning.
\newblock {\em Nature Machine Intelligence}, pages 1--13, 2022.

\bibitem{hulora}
Edward~J Hu, Phillip Wallis, Zeyuan Allen-Zhu, Yuanzhi Li, Shean Wang, Lu~Wang,
  Weizhu Chen, et~al.
\newblock Lora: Low-rank adaptation of large language models.
\newblock In {\em International Conference on Learning Representations}.

\bibitem{castro2018end}
Francisco~M Castro, Manuel~J Mar{\'\i}n-Jim{\'e}nez, et~al.
\newblock End-to-end incremental learning.
\newblock In {\em Proceedings of the European Conference on Computer Vision},
  pages 233--248, 2018.

\bibitem{Hou2019LearningAU}
Saihui Hou, Xinyu Pan, Chen~Change Loy, Zilei Wang, and D.~Lin.
\newblock Learning a unified classifier incrementally via rebalancing.
\newblock In {\em Proceedings of the IEEE/CVF Conference on Computer Vision and
  Pattern Recognition}, pages 831--839, 2019.

\bibitem{ahn2021ss}
Hongjoon Ahn, Jihwan Kwak, Subin Lim, Hyeonsu Bang, Hyojun Kim, and Taesup
  Moon.
\newblock Ss-il: Separated softmax for incremental learning.
\newblock In {\em Proceedings of the IEEE/CVF International Conference on
  Computer Vision}, pages 844--853, 2021.

\bibitem{lee2019overcoming}
Kibok Lee, Kimin Lee, Jinwoo Shin, and Honglak Lee.
\newblock Overcoming catastrophic forgetting with unlabeled data in the wild.
\newblock In {\em Proceedings of the IEEE/CVF International Conference on
  Computer Vision}, pages 312--321, 2019.

\bibitem{zhou2021co}
Da-Wei Zhou, Han-Jia Ye, and De-Chuan Zhan.
\newblock Co-transport for class-incremental learning.
\newblock In {\em ACM Multimedia Conference}, pages 1645--1654, 2021.

\bibitem{shi2022mimicking}
Yujun Shi, Kuangqi Zhou, Jian Liang, Zihang Jiang, Jiashi Feng, Philip~HS Torr,
  Song Bai, and Vincent~YF Tan.
\newblock Mimicking the oracle: an initial phase decorrelation approach for
  class incremental learning.
\newblock In {\em Proceedings of the IEEE/CVF Conference on Computer Vision and
  Pattern Recognition}, pages 16722--16731, 2022.

\bibitem{tang2022learning}
Yu-Ming Tang, Yi-Xing Peng, and Wei-Shi Zheng.
\newblock Learning to imagine: Diversify memory for incremental learning using
  unlabeled data.
\newblock In {\em Proceedings of the IEEE/CVF Conference on Computer Vision and
  Pattern Recognition}, pages 9549--9558, 2022.

\bibitem{kang2022class}
Minsoo Kang, Jaeyoo Park, and Bohyung Han.
\newblock Class-incremental learning by knowledge distillation with adaptive
  feature consolidation.
\newblock In {\em Proceedings of the IEEE/CVF Conference on Computer Vision and
  Pattern Recognition}, pages 16071--16080, 2022.

\bibitem{Wu2019LargeSI}
Y.~Wu, Yan-Jia Chen, Lijuan Wang, et~al.
\newblock Large scale incremental learning.
\newblock In {\em Proceedings of the IEEE/CVF Conference on Computer Vision and
  Pattern Recognition}, pages 374--382, 2019.

\bibitem{belouadah2019il2m}
Eden Belouadah and Adrian Popescu.
\newblock Il2m: Class incremental learning with dual memory.
\newblock In {\em Proceedings of the IEEE/CVF International Conference on
  Computer Vision}, pages 583--592, 2019.

\bibitem{ZhaoXGZX20}
Bowen Zhao, Xi~Xiao, Guojun Gan, Bin Zhang, and Shu{-}Tao Xia.
\newblock Maintaining discrimination and fairness in class incremental
  learning.
\newblock In {\em Proceedings of the IEEE/CVF Conference on Computer Vision and
  Pattern Recognition}, pages 13205--13214, 2020.

\bibitem{belouadah2020scail}
Eden Belouadah and Adrian Popescu.
\newblock Scail: Classifier weights scaling for class incremental learning.
\newblock In {\em IEEE/CVF Winter Conference on Applications of Computer
  Vision}, pages 1266--1275, 2020.

\bibitem{slim2022dataset}
Habib Slim, Eden Belouadah, Adrian Popescu, and Darian Onchis.
\newblock Dataset knowledge transfer for class-incremental learning without
  memory.
\newblock In {\em IEEE/CVF Winter Conference on Applications of Computer
  Vision}, pages 483--492, 2022.

\bibitem{zhu2023imitating}
Fei Zhu, Zhen Cheng, Xu-Yao Zhang, and Cheng-Lin Liu.
\newblock Imitating the oracle: Towards calibrated model for class incremental
  learning.
\newblock {\em Neural Networks}, 164:38--48, 2023.

\bibitem{liu2021rmm}
Yaoyao Liu, Bernt Schiele, and Qianru Sun.
\newblock Rmm: Reinforced memory management for class-incremental learning.
\newblock {\em Advances in Neural Information Processing Systems}, 34, 2021.

\bibitem{smith2021always}
James Smith, Yen-Chang Hsu, Jonathan Balloch, Yilin Shen, Hongxia Jin, and
  Zsolt Kira.
\newblock Always be dreaming: A new approach for data-free class-incremental
  learning.
\newblock In {\em Proceedings of the IEEE/CVF International Conference on
  Computer Vision}, pages 9374--9384, 2021.

\bibitem{KemkerK18}
Ronald Kemker and Christopher Kanan.
\newblock Fearnet: Brain-inspired model for incremental learning.
\newblock In {\em International Conference on Learning Representations}, 2018.

\bibitem{gao2023ddgr}
Rui Gao and Weiwei Liu.
\newblock Ddgr: Continual learning with deep diffusion-based generative replay.
\newblock In {\em International Conference on Machine Learning}, 2023.

\bibitem{boschini2022class}
Matteo Boschini, Lorenzo Bonicelli, Pietro Buzzega, Angelo Porrello, and Simone
  Calderara.
\newblock Class-incremental continual learning into the extended der-verse.
\newblock {\em IEEE Transactions on Pattern Analysis and Machine Intelligence},
  2022.

\bibitem{tao2020topology}
Xiaoyu Tao, Xinyuan Chang, Xiaopeng Hong, Xing Wei, and Yihong Gong.
\newblock Topology-preserving class-incremental learning.
\newblock In {\em Proceedings of the European Conference on Computer Vision},
  pages 254--270, 2020.

\bibitem{simon2021learning}
Christian Simon, Piotr Koniusz, and Mehrtash Harandi.
\newblock On learning the geodesic path for incremental learning.
\newblock In {\em Proceedings of the IEEE/CVF Conference on Computer Vision and
  Pattern Recognition}, pages 1591--1600, 2021.

\bibitem{hu2021distilling}
Xinting Hu, Kaihua Tang, Chunyan Miao, Xian-Sheng Hua, and Hanwang Zhang.
\newblock Distilling causal effect of data in class-incremental learning.
\newblock In {\em Proceedings of the IEEE/CVF Conference on Computer Vision and
  Pattern Recognition}, pages 3957--3966, 2021.

\bibitem{cha2021co2l}
Hyuntak Cha, Jaeho Lee, and Jinwoo Shin.
\newblock Co2l: Contrastive continual learning.
\newblock In {\em Proceedings of the IEEE/CVF International Conference on
  Computer Vision}, pages 9516--9525, 2021.

\bibitem{ashok2022class}
Arjun Ashok, KJ~Joseph, and Vineeth~N Balasubramanian.
\newblock Class-incremental learning with cross-space clustering and controlled
  transfer.
\newblock In {\em Proceedings of the European Conference on Computer Vision},
  pages 105--122. Springer, 2022.

\bibitem{rusu2016progressive}
Andrei~A Rusu, Neil~C Rabinowitz, Guillaume Desjardins, et~al.
\newblock Progressive neural networks.
\newblock {\em arXiv preprint arXiv:1606.04671}, 2016.

\bibitem{mallya2018packnet}
Arun Mallya and Svetlana Lazebnik.
\newblock Packnet: Adding multiple tasks to a single network by iterative
  pruning.
\newblock In {\em Proceedings of the IEEE/CVF Conference on Computer Vision and
  Pattern Recognition}, pages 7765--7773, 2018.

\bibitem{kim2021split}
Jong-Yeong Kim and Dong-Wan Choi.
\newblock Split-and-bridge: Adaptable class incremental learning within a
  single neural network.
\newblock In {\em Proceedings of the AAAI Conference on Artificial
  Intelligence}, volume~35, pages 8137--8145, 2021.

\bibitem{pham2021dualnet}
Quang Pham, Chenghao Liu, and Steven Hoi.
\newblock Dualnet: Continual learning, fast and slow.
\newblock {\em Advances in Neural Information Processing Systems}, 34, 2021.

\bibitem{gao2023dkt}
Xinyuan Gao, Yuhang He, Songlin Dong, Jie Cheng, Xing Wei, and Yihong Gong.
\newblock Dkt: Diverse knowledge transfer transformer for class incremental
  learning.
\newblock In {\em Proceedings of the IEEE/CVF Conference on Computer Vision and
  Pattern Recognition}, pages 24236--24245, 2023.

\bibitem{hu2023dense}
Zhiyuan Hu, Yunsheng Li, Jiancheng Lyu, Dashan Gao, and Nuno Vasconcelos.
\newblock Dense network expansion for class incremental learning.
\newblock In {\em Proceedings of the IEEE/CVF Conference on Computer Vision and
  Pattern Recognition}, pages 11858--11867, 2023.

\bibitem{zeng2019continual}
Guanxiong Zeng, Yang Chen, Bo~Cui, and Shan Yu.
\newblock Continual learning of context-dependent processing in neural
  networks.
\newblock {\em Nature Machine Intelligence}, 1(8):364--372, 2019.

\bibitem{Yu2020SemanticDC}
Lu~Yu, Bartlomiej Twardowski, X.~Liu, L.~Herranz, et~al.
\newblock Semantic drift compensation for class-incremental learning.
\newblock {\em Proceedings of the IEEE/CVF Conference on Computer Vision and
  Pattern Recognition}, pages 6980--6989, 2020.

\bibitem{DharSPWC19}
Prithviraj Dhar, Rajat~Vikram Singh, Kuan{-}Chuan Peng, et~al.
\newblock Learning without memorizing.
\newblock In {\em Proceedings of the IEEE/CVF Conference on Computer Vision and
  Pattern Recognition}, pages 5138--5146, 2019.

\bibitem{zhu2021calibration}
Fei Zhu, Xu-Yao Zhang, and Cheng-Lin Liu.
\newblock Calibration for non-exemplar based class-incremental learning.
\newblock In {\em International Conference on Multimedia and Expo}, pages 1--6,
  2021.

\bibitem{magistrielastic}
Simone Magistri, Tomaso Trinci, Albin Soutif, Joost van~de Weijer, and Andrew~D
  Bagdanov.
\newblock Elastic feature consolidation for cold start exemplar-free
  incremental learning.
\newblock In {\em The Twelfth International Conference on Learning
  Representations}.

\bibitem{wang2022learning}
Zifeng Wang, Zizhao Zhang, Chen-Yu Lee, Han Zhang, Ruoxi Sun, Xiaoqi Ren,
  Guolong Su, Vincent Perot, Jennifer Dy, and Tomas Pfister.
\newblock Learning to prompt for continual learning.
\newblock In {\em Proceedings of the IEEE/CVF Conference on Computer Vision and
  Pattern Recognition}, pages 139--149, 2022.

\bibitem{wang2022dualprompt}
Zifeng Wang, Zizhao Zhang, Sayna Ebrahimi, Ruoxi Sun, Han Zhang, Chen-Yu Lee,
  Xiaoqi Ren, Guolong Su, Vincent Perot, Jennifer Dy, et~al.
\newblock Dualprompt: Complementary prompting for rehearsal-free continual
  learning.
\newblock In {\em Proceedings of the European Conference on Computer Vision},
  pages 631--648. Springer, 2022.

\bibitem{smith2022coda}
James~Seale Smith, Leonid Karlinsky, Vyshnavi Gutta, Paola Cascante-Bonilla,
  Donghyun Kim, Assaf Arbelle, Rameswar Panda, Rogerio Feris, and Zsolt Kira.
\newblock Coda-prompt: Continual decomposed attention-based prompting for
  rehearsal-free continual learning.
\newblock In {\em Proceedings of the IEEE/CVF Conference on Computer Vision and
  Pattern Recognition}, 2023.

\bibitem{villa2022pivot}
Andr{\'e}s Villa, Juan~Le{\'o}n Alc{\'a}zar, Motasem Alfarra, Kumail Alhamoud,
  Julio Hurtado, Fabian~Caba Heilbron, Alvaro Soto, and Bernard Ghanem.
\newblock Pivot: Prompting for video continual learning.
\newblock In {\em Proceedings of the IEEE/CVF Conference on Computer Vision and
  Pattern Recognition}, 2023.

\bibitem{zhou2023revisiting}
Da-Wei Zhou, Han-Jia Ye, De-Chuan Zhan, and Ziwei Liu.
\newblock Revisiting class-incremental learning with pre-trained models:
  Generalizability and adaptivity are all you need.
\newblock {\em arXiv preprint arXiv:2303.07338}, 2023.

\bibitem{liu2023class}
Xialei Liu, Xusheng Cao, Haori Lu, Jia-wen Xiao, Andrew~D Bagdanov, and
  Ming-Ming Cheng.
\newblock Class incremental learning with pre-trained vision-language models.
\newblock {\em arXiv preprint arXiv:2310.20348}, 2023.

\bibitem{cao2024generative}
Xusheng Cao, Haori Lu, Linlan Huang, Xialei Liu, and Ming-Ming Cheng.
\newblock Generative multi-modal models are good class incremental learners.
\newblock In {\em Proceedings of the IEEE/CVF Conference on Computer Vision and
  Pattern Recognition}, pages 28706--28717, 2024.

\bibitem{kim2022multi}
Gyuhak Kim, Bing Liu, and Zixuan Ke.
\newblock A multi-head model for continual learning via out-of-distribution
  replay.
\newblock In {\em Conference on Lifelong Learning Agents}, pages 548--563.
  PMLR, 2022.

\bibitem{mirzadeh2020understanding}
Seyed~Iman Mirzadeh, Mehrdad Farajtabar, Razvan Pascanu, and Hassan
  Ghasemzadeh.
\newblock Understanding the role of training regimes in continual learning.
\newblock In {\em Advances in Neural Information Processing Systems}, 2020.

\bibitem{de2022continual}
Matthias De~Lange, Gido van~de Ven, and Tinne Tuytelaars.
\newblock Continual evaluation for lifelong learning: Identifying the stability
  gap.
\newblock In {\em International Conference on Learning Representations}, 2023.

\bibitem{mirzadeh2022wide}
Seyed~Iman Mirzadeh, Arslan Chaudhry, Dong Yin, Huiyi Hu, Razvan Pascanu, Dilan
  Gorur, and Mehrdad Farajtabar.
\newblock Wide neural networks forget less catastrophically.
\newblock In {\em International Conference on Machine Learning}, pages
  15699--15717, 2022.

\bibitem{kimtheoretical}
Gyuhak Kim, Changnan Xiao, Tatsuya Konishi, Zixuan Ke, and Bing Liu.
\newblock A theoretical study on solving continual learning.
\newblock In {\em Advances in Neural Information Processing Systems}, 2022.

\bibitem{zhuacta}
Fei Zhu, Xu-Yao Zhang, and Cheng-Lin Liu.
\newblock Class incremental learning: A review and performance evaluation.
\newblock {\em Acta Automatica Sinica}, 49(3):635--660, 2023.

\bibitem{verwimp2021rehearsal}
Eli Verwimp, Matthias De~Lange, and Tinne Tuytelaars.
\newblock Rehearsal revealed: The limits and merits of revisiting samples in
  continual learning.
\newblock In {\em Proceedings of the IEEE/CVF International Conference on
  Computer Vision}, pages 9385--9394, 2021.

\bibitem{lee2021continual}
Sebastian Lee, Sebastian Goldt, and Andrew Saxe.
\newblock Continual learning in the teacher-student setup: Impact of task
  similarity.
\newblock In {\em International Conference on Machine Learning}, pages
  6109--6119, 2021.

\bibitem{cheng2023average}
Zhen Cheng, Fei Zhu, Xu-Yao Zhang, and Cheng-Lin Liu.
\newblock Average of pruning: Improving performance and stability of
  out-of-distribution detection.
\newblock {\em arXiv preprint arXiv:2303.01201}, 2023.

\bibitem{liu2022long}
Xialei Liu, Yu-Song Hu, Xu-Sheng Cao, Andrew~D Bagdanov, Ke~Li, and Ming-Ming
  Cheng.
\newblock Long-tailed class incremental learning.
\newblock In {\em Proceedings of the European Conference on Computer Vision},
  pages 495--512. Springer, 2022.

\bibitem{zhao2024continual}
Hongbo Zhao, Bolin Ni, Junsong Fan, Yuxi Wang, Yuntao Chen, Gaofeng Meng, and
  Zhaoxiang Zhang.
\newblock Continual forgetting for pre-trained vision models.
\newblock In {\em Proceedings of the IEEE/CVF Conference on Computer Vision and
  Pattern Recognition}, pages 28631--28642, 2024.

\bibitem{fini2022self}
Enrico Fini, Victor G~Turrisi Da~Costa, Xavier Alameda-Pineda, Elisa Ricci,
  Karteek Alahari, and Julien Mairal.
\newblock Self-supervised models are continual learners.
\newblock In {\em Proceedings of the IEEE/CVF Conference on Computer Vision and
  Pattern Recognition}, pages 9621--9630, 2022.

\bibitem{liu2024branch}
Wenzhuo Liu, Fei Zhu, and Cheng-Lin Liu.
\newblock Branch-tuning: Balancing stability and plasticity for continual
  self-supervised learning.
\newblock {\em arXiv preprint arXiv:2403.18266}, 2024.

\bibitem{guo2024federated}
Haiyang Guo, Fei Zhu, Wenzhuo Liu, Xu-Yao Zhang, and Cheng-Lin Liu.
\newblock Federated class-incremental learning with prototype guided
  transformer.
\newblock {\em arXiv preprint arXiv:2401.02094}, 2024.

\bibitem{liu2023learnable}
Binghao Liu, Boyu Yang, Lingxi Xie, Ren Wang, Qi~Tian, and Qixiang Ye.
\newblock Learnable distribution calibration for few-shot class-incremental
  learning.
\newblock {\em IEEE Transactions on Pattern Analysis and Machine Intelligence},
  2023.

\bibitem{yang2022dynamic}
Boyu Yang, Mingbao Lin, Yunxiao Zhang, Binghao Liu, Xiaodan Liang, Rongrong Ji,
  and Qixiang Ye.
\newblock Dynamic support network for few-shot class incremental learning.
\newblock {\em IEEE Transactions on Pattern Analysis and Machine Intelligence},
  45(3):2945--2951, 2022.

\bibitem{Aljundi2019ContinualLI}
Rahaf Aljundi.
\newblock Continual learning in neural networks.
\newblock {\em ArXiv}, abs/1910.02718, 2019.

\bibitem{hinton2015distilling}
Geoffrey Hinton, Oriol Vinyals, Jeff Dean, et~al.
\newblock Distilling the knowledge in a neural network.
\newblock {\em arXiv preprint arXiv:1503.02531}, 2(7), 2015.

\bibitem{aljundi2018memory}
Rahaf Aljundi, Francesca Babiloni, et~al.
\newblock Memory aware synapses: Learning what (not) to forget.
\newblock In {\em Proceedings of the European Conference on Computer Vision},
  pages 139--154, 2018.

\bibitem{Gidaris2018UnsupervisedRL}
Spyros Gidaris, Praveer Singh, and Nikos Komodakis.
\newblock Unsupervised representation learning by predicting image rotations.
\newblock In {\em International Conference on Learning Representations 2018},
  2018.

\bibitem{lee20_sla}
Hankook Lee, Sung~Ju Hwang, and Jinwoo Shin.
\newblock Self-supervised label augmentation via input transformations.
\newblock In {\em International Conference on Machine Learning}, 2020.

\bibitem{maaten2013learning}
Laurens Maaten, Minmin Chen, Stephen Tyree, and Kilian Weinberger.
\newblock Learning with marginalized corrupted features.
\newblock In {\em International Conference on Machine Learning}, pages
  410--418, 2013.

\bibitem{wang2021regularizing}
Yulin Wang, Gao Huang, Shiji Song, et~al.
\newblock Regularizing deep networks with semantic data augmentation.
\newblock {\em IEEE Transactions on Pattern Analysis and Machine Intelligence},
  2021.

\bibitem{ganaie2022ensemble}
Mudasir~A Ganaie, Minghui Hu, AK~Malik, M~Tanveer, and PN~Suganthan.
\newblock Ensemble deep learning: A review.
\newblock {\em Engineering Applications of Artificial Intelligence},
  115:105151, 2022.

\bibitem{van2008visualizing}
Laurens Van~der Maaten and Geoffrey Hinton.
\newblock Visualizing data using t-sne.
\newblock {\em Journal of machine learning research}, 9(11), 2008.

\bibitem{LeCun2005TheMD}
Yann LeCun.
\newblock The mnist database of handwritten digits.
\newblock {\em http://yann. lecun. com/exdb/mnist/}, 1998.

\bibitem{dosovitskiy2020image}
Alexey Dosovitskiy, Lucas Beyer, Alexander Kolesnikov, Dirk Weissenborn,
  Xiaohua Zhai, Thomas Unterthiner, Mostafa Dehghani, Matthias Minderer, Georg
  Heigold, Sylvain Gelly, et~al.
\newblock An image is worth 16x16 words: Transformers for image recognition at
  scale.
\newblock In {\em International Conference on Learning Representations}, 2020.

\bibitem{he2016deep}
Kaiming He, Xiangyu Zhang, Shaoqing Ren, and Jian Sun.
\newblock Deep residual learning for image recognition.
\newblock In {\em Proceedings of the IEEE/CVF Conference on Computer Vision and
  Pattern Recognition}, pages 770--778, 2016.

\bibitem{krizhevsky2009learning}
Alex Krizhevsky, Geoffrey Hinton, et~al.
\newblock Learning multiple layers of features from tiny images.
\newblock Technical report, Citeseer, 2009.

\bibitem{LeeLLS18}
Kimin Lee, Kibok Lee, Honglak Lee, and Jinwoo Shin.
\newblock A simple unified framework for detecting out-of-distribution samples
  and adversarial attacks.
\newblock In {\em Advances in Neural Information Processing Systems}, pages
  7167--7177, 2018.

\bibitem{MensinkVPC13}
Thomas Mensink, Jakob~J. Verbeek, Florent Perronnin, and Gabriela Csurka.
\newblock Distance-based image classification: Generalizing to new classes at
  near-zero cost.
\newblock {\em IEEE Transactions on Pattern Analysis and Machine Intelligence},
  pages 2624--2637, 2013.

\bibitem{kingma2015adam}
Diederik~P Kingma and Jimmy Ba.
\newblock Adam: A method for stochastic optimization.
\newblock In {\em International Conference on Learning Representations}, 2015.

\bibitem{Zhu_2021_NeurIPS}
Fei Zhu, Zhen Cheng, Xu-yao Zhang, and Cheng-lin Liu.
\newblock Class-incremental learning via dual augmentation.
\newblock In {\em Advances in Neural Information Processing Systems},
  volume~34, 2021.

\bibitem{zhu2022self}
Kai Zhu, Wei Zhai, Yang Cao, Jiebo Luo, and Zheng-Jun Zha.
\newblock Self-sustaining representation expansion for non-exemplar
  class-incremental learning.
\newblock In {\em Proceedings of the IEEE/CVF Conference on Computer Vision and
  Pattern Recognition}, pages 9296--9305, 2022.

\bibitem{pouransari2015tiny}
Hadi Pouransari and Saman Ghili.
\newblock Tiny imagenet visual recognition challenge.
\newblock {\em CS231N course, Stanford Univ., Stanford, CA, USA}, 2015.

\bibitem{deng2009imagenet}
Jia Deng, Wei Dong, Richard Socher, et~al.
\newblock Imagenet: A large-scale hierarchical image database.
\newblock In {\em Proceedings of the IEEE/CVF Conference on Computer Vision and
  Pattern Recognition}, pages 248--255, 2009.

\bibitem{liu2020mnemonics}
Yaoyao Liu, Yuting Su, An-An Liu, Bernt Schiele, and Qianru Sun.
\newblock Mnemonics training: Multi-class incremental learning without
  forgetting.
\newblock In {\em Proceedings of the IEEE/CVF conference on Computer Vision and
  Pattern Recognition}, pages 12245--12254, 2020.

\bibitem{touvron2021training}
Hugo Touvron, Matthieu Cord, Matthijs Douze, Francisco Massa, Alexandre
  Sablayrolles, and Herv{\'e} J{\'e}gou.
\newblock Training data-efficient image transformers \& distillation through
  attention.
\newblock In {\em International conference on machine learning}, pages
  10347--10357. PMLR, 2021.

\bibitem{hayes2020lifelong}
Tyler~L Hayes and Christopher Kanan.
\newblock Lifelong machine learning with deep streaming linear discriminant
  analysis.
\newblock In {\em Proceedings of the IEEE/CVF conference on computer vision and
  pattern recognition workshops}, pages 220--221, 2020.

\bibitem{chaudhry2018efficient}
Arslan Chaudhry, Marc’Aurelio Ranzato, Marcus Rohrbach, and Mohamed
  Elhoseiny.
\newblock Efficient lifelong learning with a-gem.
\newblock In {\em International Conference on Learning Representations}, 2018.

\bibitem{buzzega2020dark}
Pietro Buzzega, Matteo Boschini, et~al.
\newblock Dark experience for general continual learning: a strong, simple
  baseline.
\newblock In {\em Advances in Neural Information Processing Systems}, 2020.

\bibitem{chaudhry2021using}
Arslan Chaudhry, Albert Gordo, Puneet Dokania, Philip Torr, and David
  Lopez-Paz.
\newblock Using hindsight to anchor past knowledge in continual learning.
\newblock In {\em Proceedings of the AAAI conference on artificial
  intelligence}, volume~35, pages 6993--7001, 2021.

\bibitem{wang2022beef}
Fu-Yun Wang, Da-Wei Zhou, Liu Liu, Han-Jia Ye, Yatao Bian, De-Chuan Zhan, and
  Peilin Zhao.
\newblock Beef: Bi-compatible class-incremental learning via energy-based
  expansion and fusion.
\newblock In {\em The eleventh international conference on learning
  representations}, 2022.

\bibitem{welling2009herding}
Max Welling.
\newblock Herding dynamical weights to learn.
\newblock In {\em International Conference on Machine Learning}, pages
  1121--1128, 2009.

\bibitem{Hsu2018ReevaluatingCL}
Yen-Chang Hsu, Y.~Liu, and Z.~Kira.
\newblock Re-evaluating continual learning scenarios: A categorization and case
  for strong baselines.
\newblock {\em ArXiv}, abs/1810.12488, 2018.

\bibitem{Ven2019ThreeSF}
Gido~M. van~de Ven and A.~Tolias.
\newblock Three scenarios for continual learning.
\newblock {\em ArXiv}, abs/1904.07734, 2019.

\bibitem{yin2020dreaming}
Hongxu Yin, Pavlo Molchanov, Jose~M Alvarez, Zhizhong Li, Arun Mallya, Derek
  Hoiem, Niraj~K Jha, and Jan Kautz.
\newblock Dreaming to distill: Data-free knowledge transfer via deepinversion.
\newblock In {\em Proceedings of the IEEE/CVF Conference on Computer Vision and
  Pattern Recognition}, pages 8715--8724, 2020.

\bibitem{li2024esdb}
Kunchi Li, Hongyang Chen, Jun Wan, and Shan Yu.
\newblock Esdb: Expand the shrinking decision boundary via one-to-many
  information matching for continual learning with small memory.
\newblock {\em IEEE Transactions on Circuits and Systems for Video Technology},
  2024.

\bibitem{guo2017calibration}
Chuan Guo, Geoff Pleiss, Yu~Sun, and Kilian~Q Weinberger.
\newblock On calibration of modern neural networks.
\newblock In {\em International conference on machine learning}, pages
  1321--1330. PMLR, 2017.

\bibitem{Naeini2015ObtainingWC}
Mahdi~Pakdaman Naeini, Gregory~F. Cooper, and Milos Hauskrecht.
\newblock Obtaining well calibrated probabilities using bayesian binning.
\newblock In {\em AAAI}, pages 2901--2907, 2015.

\bibitem{kim2023stability}
Dongwan Kim and Bohyung Han.
\newblock On the stability-plasticity dilemma of class-incremental learning.
\newblock In {\em Proceedings of the IEEE/CVF Conference on Computer Vision and
  Pattern Recognition}, pages 20196--20204, 2023.

\bibitem{zhu2022rethinking}
Fei Zhu, Zhen Cheng, Xu-Yao Zhang, and Cheng-Lin Liu.
\newblock Rethinking confidence calibration for failure prediction.
\newblock In {\em European Conference on Computer Vision}, pages 518--536,
  2022.

\bibitem{hendrycksbenchmarking}
Dan Hendrycks and Thomas Dietterich.
\newblock Benchmarking neural network robustness to common corruptions and
  perturbations.
\newblock In {\em International Conference on Learning Representations}, 2019.

\bibitem{luo2021no}
Mi~Luo, Fei Chen, Dapeng Hu, Yifan Zhang, Jian Liang, and Jiashi Feng.
\newblock No fear of heterogeneity: Classifier calibration for federated
  learning with non-iid data.
\newblock In {\em Advances in Neural Information Processing Systems}, 2021.

\bibitem{lesort2020continual}
Timoth{\'e}e Lesort, Vincenzo Lomonaco, Andrei Stoian, Davide Maltoni, David
  Filliat, and Natalia D{\'\i}az-Rodr{\'\i}guez.
\newblock Continual learning for robotics: Definition, framework, learning
  strategies, opportunities and challenges.
\newblock {\em Information Fusion}, 58:52--68, 2020.

\bibitem{biederman1985human}
Irving Biederman.
\newblock Human image understanding: Recent research and a theory.
\newblock {\em Computer Vision, Graphics, and Image Processing}, 32(1):29--73,
  1985.

\end{thebibliography}
}

\end{document}